# GPRInvNet: Deep Learning-Based Ground-Penetrating Radar Data Inversion for Tunnel Linings

Bin Liu⬡, Yuxiao Ren⬡, Hanchi Liu, Hui Xu, Zhengfang Wang⬡, Anthony G. Cohn⬡, and Peng Jiang⬡, *Member, IEEE*

*Abstract*—A DNN architecture referred to as GPRInvNet was proposed to tackle the challenges of mapping the ground-penetrating radar (GPR) B-Scan data to complex permittivity maps of subsurface structures. The GPRInvNet consisted of a trace-to-trace encoder and a decoder. It was specially designed to take into account the characteristics of GPR inversion when faced with complex GPR B-Scan data, as well as addressing the spatial alignment issues between time-series B-Scan data and spatial permittivity maps. It displayed the ability to fuse features from several adjacent traces on the B-Scan data to enhance each trace, and then further condense the features of each trace separately. As a result, the sensitive zones on the permittivity maps spatially aligned to the enhanced trace could be reconstructed accurately. The GPRInvNet has been utilized to reconstruct the permittivity map of tunnel linings. A diverse range of dielectric models of tunnel linings containing complex defects has been reconstructed using GPRInvNet. The results have demonstrated that the GPRInvNet is capable of effectively reconstructing complex tunnel lining defects with clear boundaries. Comparative results with existing baseline methods also demonstrated the superiority of the GPRInvNet. For the purpose of generalizing the GPRInvNet to real GPR data, some background noise patches recorded from practical model testing were integrated into the synthetic GPR data to retrain the GPRInvNet. The model testing has been conducted for validation, and experimental results revealed that the GPRInvNet had also achieved satisfactory results with regard to the real data.

## I. INTRODUCTION

G ROUND-PENETRATING radar (GPR) has been extensively used in many applications, including in the fields of glaciology, archeology, and civil and geotechnical engineering. For example, it has been utilized in geological surveys, buried object detections, and the detections of subsurface structures [1]–[3]. Among the aforementioned applications, the nondestructive inspections of tunnel lining structures have been popular [4], [5]. The GPR transmits an electromagnetic wave into the tunnel lining structure and receives echoes to form B-scan images, from which the structural conditions of the tunnel linings can be deduced [6], [7]. The inspection of the structural condition of the tunnel lining is of major importance to the safe operation of tunnels [8]. However, due to various geological and environmental factors, aging, increased loading, man-made impacts, and irregular construction, tunnel linings progressively deteriorate, leading to many defects, including lining voids, cracks, delamination, lining leakage, and non-compactness of concrete. These covert defects, which are generally located inside the tunnel lining, may reduce the bearing capacity of the lining, as well as affecting the normal operations of tunnels, shortening tunnel durability, or even inducing safety incidents [9], [10]. Several incidents have occurred due to the deterioration of the tunnel lining structures, such as the Big Dig ceiling collapse in Boston (2006) and the Sasago Tunnel collapse in Tokyo (2012) [11].

The translation of the electromagnetic information, which is stored in the B-Scan into inner defect-related information (such as locations, shapes, and dielectric properties), is of major importance in tunnel lining defect inspection. There are a number of existing methods for GPR inversion which focus on mapping the dielectric distributions of the structures to be detected based on the recorded GPR data [12]. These methods mainly include common-midpoint velocity analysis [13], ray-based methods [14], reverse-time migration

Manuscript received December 25, 2019; revised March 2, 2020, April 30, 2020, August 10, 2020, and October 30, 2020; accepted December 12, 2020. Date of publication January 13, 2021; date of current version September 27, 2021. This work was supported in part by the Joint Research Fund of National Natural Science Foundation of China under Grant U1806226; in part by the Key Project of National Natural Science Foundation of China under Grant 51739007; in part by the National Science Fund for Outstanding Young Scholars under Grant 51922067; and in part by the National Natural Science Foundation of China under Grant 61702301, and Grant 41877230. *(Corresponding authors: Zhengfang Wang; Peng Jiang.)*

Bin Liu is with the School of Qilu Transportation, Shandong University, Jinan 250061, China, also with the Geotechnical and Structural Engineering Techniques Research Center, Shandong University, Jinan 250061, China, and also with the Data Science Institute, Shandong University, Jinan 250061, China (e-mail: liubin0635@163.com).

Yuxiao Ren is with the School of Qilu Transportation, Shandong University, Jinan 250061, China, and also with the Geotechnical and Structural Engineering Techniques Research Center, Shandong University, Jinan 250061, China (e-mail: ryxchina@gmail.com).

Hanchi Liu and Zhengfang Wang are with the School of Control Science and Engineering, Shandong University, Jinan 250061, China (e-mail: 201934495@mail.sdu.edu.cn; wangzhengfangsdu@hotmail.com).

Hui Xu is with the Geotechnical and Structural Engineering Techniques Research Center, Shandong University, Jinan 250061, China (e-mail: 1162259518@qq.com).

Anthony G. Cohn is with the School of Computing, University of Leeds, Leeds LS2 9JT, U.K. (e-mail: a.g.cohn@leeds.ac.uk).

Peng Jiang is with the School of Qilu Transportation, Shandong University, Jinan 250061, China (e-mail: sdujump@gmail.com).

Digital Object Identifier 10.1109/TGRS.2020.3046454

(RTM) [15], tomography approaches [16], and full-waveform inversion (FWI) methods [17], [18]. Among these methods, the FWI is considered to be a state-of-the-art solution to qualitatively and quantitatively reconstruct images of subsurface structures. It directly employs the entire received waveforms to match with the forward modeled data. Then, it reconstructs the dielectric distributions of the structures by minimizing the misfit between the two sets of data [19], [20]. The FWI is originated in the field of seismic exploration [21] and has since been rapidly employed for processing radar data [22]. In tunnel lining-related applications, some developments of FWI have been presented to further improve performance. These include a truncated Newton method based on GPR FWI with structural constraints [23], a multiscale inversion strategy and bi-parametric FWI method [24], and a combination of improved FWI and RTM [25]. However, due to the fact that the tunnel lining defects always have irregular geometric characteristics and complex distribution patterns, the received subsurface GPR data are generally interlaced and accompanied by discontinuous and distorted echoes. Furthermore, some of the strong echoes induced by the steel rebars in tunnel linings may mask the signatures of the defects. In such cases, it has been observed that the B-scan images commonly show "pseudo-hyperbolic" morphologies or clutter [24]. Therefore, it has been found to be challenging for traditional FWI to precisely reconstruct the dielectric distributions of the targets. The locations of the lining defects may be erroneously computed, notwithstanding the considerable computational costs of the FWI methods.

In recent years, deep neural networks (DNNs) have demonstrated extraordinary abilities in applications related to image classifications [26], [27], object detections [28], semantic segmentations (pixel-level predictions) [29], [30], and image syntheses [31]. DNNs automatically learn the high-level features via the training data and then are able to estimate nonlinear mapping between the input image data and the various data domains, such as labels, text, or other images. Accordingly, some end-to-end deep learning-based inversion methods have been introduced to invert the velocity or impedance from the seismic data. For example, Das et al. [32] utilized a 1-D convolutional neural network (CNN) to predict high-resolution impedance. Araya-Polo et al. [33] proposed GeoDNN for seismic tomography. Alfarraj et al. [34] proposed a semisupervised framework for impedance inversion based on convolutional and recurrent neural networks. In regard to velocity inversion, Zhang et al. [35] developed an end-to-end framework referred to as VelocityGAN in order to reconstruct subsurface velocity directly from raw seismic waveform data. Wu et al. [36] designed Inversion-Net, which follows an auto-encoder architecture to map seismic data to a corresponding velocity model. In our previous study, we proposed the DNN-based SeisInvNet model to address weak spatial correspondence, the uncertain reflection–reception relationships between seismic data and velocity model, and the time-varying problems of seismic data [37]. It was found that the SeisInvNet could recover the details of interfaces and accurately reconstruct velocity models.

Since both GPR and seismology are wave-based geophysical techniques, these state-of-the-art methods for seismic inversions introduce new perspectives for addressing GPR inversion problems. However, it may not be an optimal choice to directly utilize the existing DNNs designed for seismic inversion in the processing of GPR data. First, due to the fact that the lining defects usually have irregular geometries and inhomogeneous distribution patterns, the rebar or complex dielectric distributions in tunnel linings may mask the effective GPR signals reflected by the defects. In such cases, the GPR data of the tunnel linings are usually more complex than the seismic data. Therefore, the DNNs for the GPR data inversion should have strong abilities to extract effective features from complex data. Second, GPR data is known to have unique spatial alignment characteristics: unlike those of seismic cases, the relationship between the reflection and reception [37] of GPR is relatively certain since there are only one transmitter and one receiver. In addition, since the electromagnetic waves of the GPR show a faster decay than those of the seismic waves, the majority of the useful signals induced by one defect will be captured in several adjacent GPR traces, rather than the traces far from the defect. Therefore, to accurately reconstruct the local details of the permittivity map using DNNs, it is more effective to make full use of the information extracted from adjacent GPR traces, rather than the global context. In this way, learning from ineffective information can be avoided.

So far, little progress has been made in the field of DNN-based GPR data inversion. The majority of the existing studies have adopted DNNs to process GPR B-Scan data for the detection of buried objects [38] and rebars [39], identification of subgrade defects [40], or the reconstruction of concealed crack profiles in pavements [41]. These methods have been focused on the tasks of classification or object detection in GPR B-Scan images, where the outputs are class labels or locations of the defects in the B-scan images, rather than the subsurface images of the structures. In terms of mapping GPR B-Scan images to subsurface images, to the best of our knowledge, the only study published so far was that presented by Alvarez et al. [42], who had adopted some deep learning networks for GPR image-to-image translation. Three widespread DNNs [Enc-Dec, U-Net, and generative adversarial network (GAN)] were employed to reconstruct subsurface images of concrete sewer pipes from GPR B-Scan images. Then the methods were validated using synthetic data, and the results indicated the feasibility of utilizing DNNs in mapping GPR images to subsurface images of a structure. This study successfully reconstructed subsurface images containing defects with regular geometries (triangles, circles, and rectangles). However, the dielectric properties of the structure were not reconstructed.

In order to accurately invert the dielectric properties of tunnel linings and reconstruct complex defects with irregular geometries, an end-to-end DNN framework was proposed in this study. The proposed framework was referred to as GPRInvNet, which consisted of a specially designed "trace-to-trace" encoding process and a decoding process. The encoder enhanced the features of each GPR trace using the information extracted from its adjacent traces. Then, the features of each trace were condensed one-by-one to generate a group of features that spatially corresponded to its own sensitive zone on the permittivity map. By doing so, the effective features could be extracted from complex B-Scan data and the spatial alignment between the input and output was retained. The GPRInvNet was first validated on a synthetic GPR data set. A diverse range of dielectric models containing complex defects was reconstructed using GPRInvNet, and a comprehensive comparative analysis was then performed. Furthermore, in order to apply the proposed model to real GPR data, some real background noise patches were integrated into the synthetic GPR data to train the GPRInvNet. The experimental results demonstrated that our method provided good results for real GPR data.

The main contributions of this study are as follows.
1) The GPRInvNet framework was proposed to accurately invert dielectric images directly from GPR data. To the best of our knowledge, this was the first deep learning-based network specifically designed for GPR data inversion.
2) The designed GPRInvNet was then successfully applied to reconstruct permittivity maps of tunnel linings containing complex defects. This study's comparative validation results demonstrated the superior performance of the proposed compared with other baseline models.
3) A method for generalizing GPRInvNet to real data was also presented in this study. The experimental results of the model testing showed that the proposed method had achieved satisfactory results on real data.

## II. Related Works

### A. Seismic Inversion

DNNs have been extensively exploited in the field of seismology. Since GPR and seismology are both wave-based geophysical techniques, they share similar properties from a data processing perspective. Therefore, in this section, we introduce related works in the area of seismic inversion. Different approaches based on deep learning have been recently proposed for seismic inversion. Seismic inversions have been attempted using CNNs [32], recurrent neural networks [34], and GANs[35]. Furthermore, various DNN improvements have also been presented, such as GeoDNN [33], VelocityGAN[35], and InversionNet[36].

In our previous study, we proposed SeisInvNet [37] to accurately invert the subsurface velocity distributions from observational data collected from the ground surface. To tackle the challenges encountered when mapping the time-series seismic wave signals to spatial images, the SeisInvNet utilized convolutional layers to encode the observational setup, neighborhood information, and global context of a single-shot seismic profile into one single seismic trace, which formed an embedding vector. Then, each embedding vector, which contained a variety of seismic information, was fused via fully connected layers to form a spatially aligned feature map of different seismic traces. The velocity model was reconstructed from all feature maps by a CNN called the velocity model decoder. These DNN-based methods provided new perspectives for GPR inversion problems.

### B. DNNs for the GPR Image to Subsurface Image Transformations

So far, studies, which have employed DNNs to map GPR B-Scan images to subsurface images, have been rare. To the best of our knowledge, the only published study to date has been the one conducted by Alvarez *et al.* [42]. In their article, three different state-of-the-art deep learning architectures were employed to reconstruct subsurface images of concrete sewer pipes from B-Scan images. The types of architecture of the three DNNs, including encoder-decoder (Enc-Dec), U-Net, and generation aggressive network (GAN), were identical to the implemented architectures previously described in [43]. The usage of different loss metrics was also evaluated and compared. The validations were conducted using synthetic data. The synthetic GPR B-Scan data and subsurface permittivity map were reformatted into image pairs. Specifically, each GPR B-Scan was reformatted to an image with the size of $125 \times 125$ pixels, which corresponded to a concrete segment with the dimensions of $250 \times 250$ mm.

The results of the comparative studies demonstrated that Enc-Dec networks using a differential structural similarity (DSSIM) loss function were able to slightly outperform U-Net and GAN for these types of GPR image-to-image transformation. The Enc-Dec network implemented in the studies consisted of an encoder and a corresponding decoder. Each convolutional layer employs a $4 \times 4$ convolutional filter, with a stride of 2 for downsampling the input image. Subsurface images containing defects with regular geometries (triangles, circles, and rectangles) were found to have been effectively reconstructed. The results had demonstrated the feasibility of utilizing DNNs to map a GPR image to a subsurface image. Although the permittivity maps of the subsurface structures had not been reconstructed in the aforementioned studies, a basis for further exploring the applications of deep learning in GPR inversions was still provided.

## III. Methodology

### A. Characteristics of DNN-Based GPR Inversion Tasks

The DNN-based inversion method is a data-driven nonlinear mapping problem [35]

$$H : P \to D. \tag{1}$$

The aim is to find the transformation that reconstructs a permittivity map of the tunnel lining structure $P^i$ from the corresponding GPR B-Scan $D^i$, where $i \in [I, N]$ ($N$ is the number of B-Scan images). Each permittivity model $P^i$ has size $[H; W]$, where $H$ represents the depth and $W$ represents the width of the permittivity model. Each GPR B-Scan $D^i$ has dimension $[T; R]$, where $T$ and $R$ denote the time step and the number of traces, respectively. The B-Scan $D^i$ consists of $R$ single GPR traces (A-Scans). In this article, $D^i r$ denotes the $r$th single trace on the B-Scan $D^i$, where $r \in [I, R]$. Generally, the single GPR trace $D^i r$ provides elapsed information along the depth of the tunnel lining. The data recorded at each time step in the single trace $D^i r$ are related to the dielectric properties at different depths. The different single traces $D^i r$, $r \in [I, R]$ correspond to the dielectric properties $P^i r*$, $r* \in [1, W]$ at different detection distances along the width direction of the permittivity map. The schematic is detailed in Fig. 1.

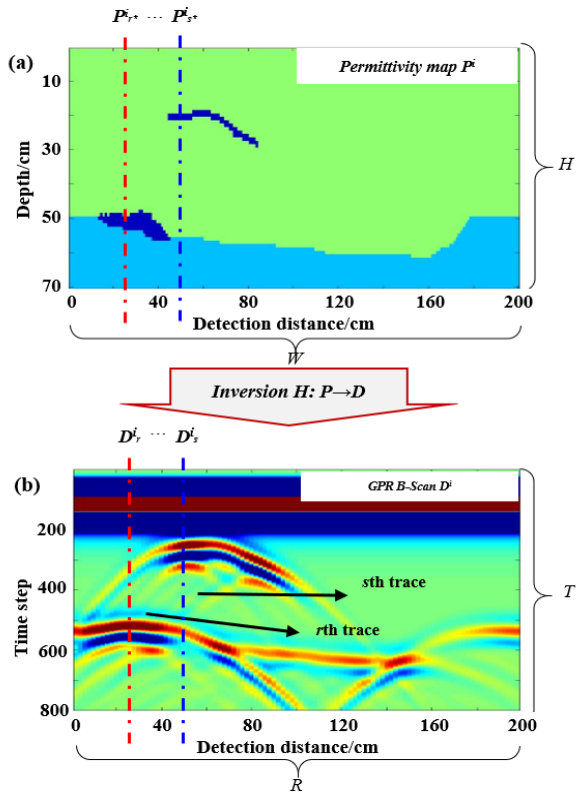

Fig. 1. Schematic of GPR inversion. (a) Permittivity map and (b) corresponding GPR B-Scan. $D_r^i$ and $D_s^i$ are the $r$th trace and the $s$th trace on the GPR B-Scan, respectively; $P_{r*}^i$ is the $r*$th column of permittivity values spatially aligned to $D_r^i$; and $P_{s*}^i$ is the $s*$th column of permittivity values spatially aligned to $D_s^i$.

The challenges of DNN-based GPR inversion are twofold as follows.

1) Due to the fact that tunnel linings may contain rebars, the dielectric distributions are usually inhomogeneous. Consequently, the GPR B-Scan data of the tunnel linings are very complex. In particular, the rebars located inside the tunnel linings may mask the effective GPR echoes from any defects, which manifests in the B-Scan as clutter, as illustrated in Fig. 2. Moreover, tunnel lining defects are known to consistently have irregular geometric characteristics, and defects with different shapes may contribute to similar B-Scan profiles under the impacts of multiple waves and scattering. In such cases, it has been found to be challenging to accurately reconstruct the details of tunnel lining defects with irregular shapes, especially those located under rebars. Therefore, a network with strong feature extracting capacity was required to make full use of the input data and to learn the effective features from the complex B-Scan images.

2) There is no specific spatial alignment between the input (GPR B-Scan) and the output (relative permittivity model) images. This is particularly significant in DNN-based inversion processes as the majority of the existing DNN methods are designed for spatially aligned data pairs. Generally, the position at which a hyperbolic echo exists on a GPR profile may not correspond to any abnormalities in the dielectric model. As can be seen in Fig. 2, the echoes induced by the multiple reflections of rebars do not align to any abnormality on the dielectric model. However, the signals induced by a defect in the dielectric model can always be observed, not only in their corresponding traces but also in several adjacent GPR traces. For example, the dielectric model at one location is not only related to the corresponding GPR trace but also to several adjacent GPR traces.

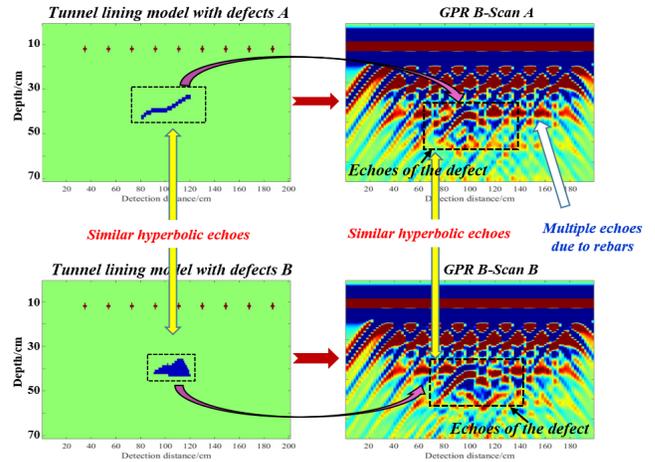

Fig. 2. Two B-Scan and tunnel lining model pairs. The tunnel lining models A and B contain two different defects under rebars; the B-Scan images are complex; and the B-Scan images of the two models are quite similar.

Various end-to-end DNN frameworks designed for image synthesis can be employed to map GPR B-Scan data to permittivity images, such as Enc-Dec, U-Net, and GAN. However, the existing networks were all originally designed for image pairs that are spatially aligned, such as photos and medical images. These networks employ fixed convolutional kernels and encode the input data into a feature vector, from which the decoder reconstructs the output [37]. However, with the increases in the dimensions of feature maps, the spatial features may be gradually lost, which may cause the details or boundaries to not be accurately reconstructed [44], [45]. For GPR data without specific spatial alignments, the existing DNNs may not be the optimal choice, and deep learning networks that explicitly consider the characteristics of GPR data may be preferable.

### B. Architecture of GPRInvNet

In this study, a novel DNN architecture for GPR inversion referred to as GPRInvNet was proposed, which displayed the ability to make full use of the information in the B-Scan and retain the spatial alignment between the input and output. The concept of the GPRInvNet framework was inspired by the previously introduced SeisInvNet model [37]. However, improvements were made to the network in order to select specific accounts of the characteristics of the GPR data. With consideration given to the complexity of GPR B-Scans of tunnel linings, the feature extraction components of the network were increased in order to extract features from the complex B-Scan data, as well as to enhance the information of each trace using the information of its adjacent traces. In addition, due to the special spatial alignment characteristics between the GPR data and the permittivity map, this study separately condensed the features of each trace, which were spatially aligned to columns of the permittivity map. Then, each column of the permittivity map was accurately reconstructed from the features of each trace. Then, the entire permittivity map could be obtained by splicing all of the slices of the permittivity map.

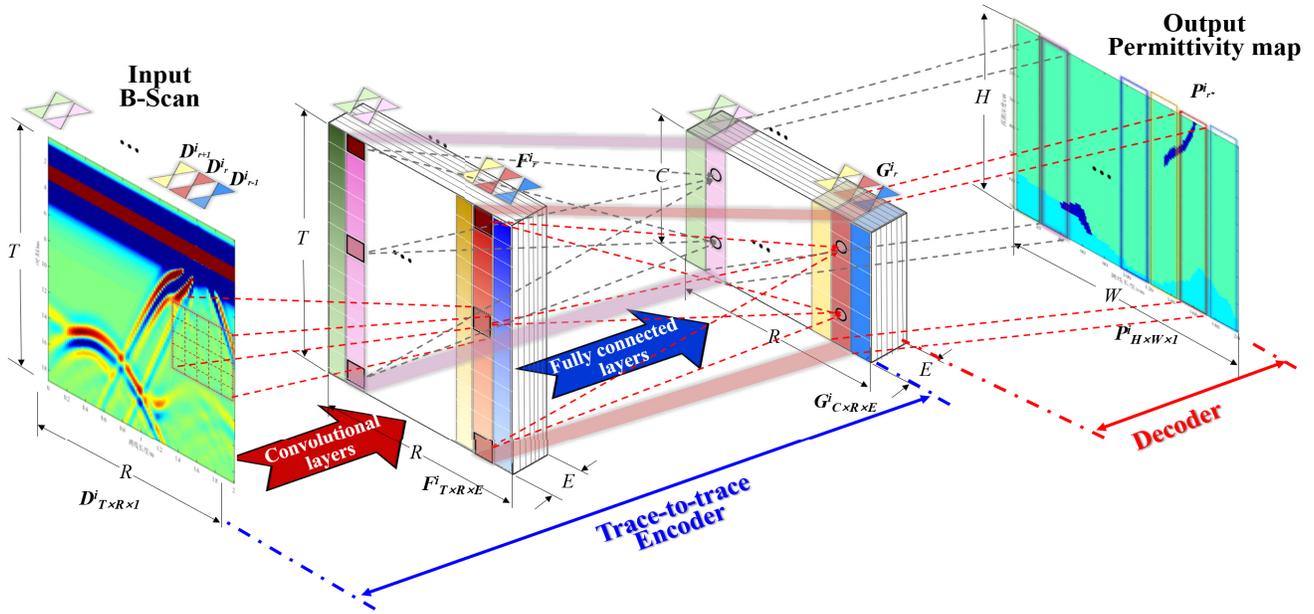

Fig. 3. Architecture of GPRInvNet. $D^i$ with a dimension of $T \times R \times 1$ is the input B-Scan data; $F^i$ with the dimension of $T \times R \times E$ indicates the feature maps after convolutional layers; a group of features $F^i_r$ spatially aligned to $D^i_r$ contains information extracted from the adjacent traces of $D^i_r$; $G^i$ with the dimension of $C \times R \times E$ is formed by separately condensing each column of $F^i$ using fully connected layers and $G^i_r$ corresponding to the compressed features of $F^i_r$; $P^i_r$ is a part of permittivity map spatially aligned to $G^i_r$; and the output of the network $P^i$ with the dimension of $H \times W \times 1$ is the reconstructed permittivity map from $D^i$.

Fig. 3 shows the architecture of the GPRInvNet framework. The GPRInvNet consisted of a specially designed encoder and a corresponding decoder. The decoder was topologically identical to the decoder component in the SeisInvNet model [37]. The key component of the GPRInvNet was its encoder, which was described as a "trace-to-trace" encoder. During the encoding process, multiple convolutional layers were first employed to enrich the information of each GPR trace without compressing its spatial dimension. Then, as the number of convolutional layers increased as per the GPR data, the capacity of feature extraction had been enhanced. This allowed the network to make full use of the information from the adjacent traces and automatically learn features from the complex B-Scan data. Moreover, since the GPR signals based on electromagnetic waves showed a trend of faster decay in amplitude when compared with the seismic waves, the useful signals observed in adjacent traces may not be detected in more remote traces. Therefore, only the effective information from the neighboring traces was fused, rather than extracting the global context from the entire B-Scan. This had effectively prevented the network from learning ineffective features.

In this article, after the features had been enhanced, several fully connected layers were utilized in the encoding process for the purpose of separately condensing the features of each trace. Then, the enhanced features of each trace were artificially aligned to a column of the permittivity map. In this study, the term "sensitive zone" refers to the column of the permittivity map which spatially aligns to the enhanced features of each trace. This operation allowed the network to accurately reconstruct the details of its sensitive zone.

During the decoding process, the high-quality sensitive zones of each trace were reconstructed. Then, all the inverted sensitive zones were spliced together to form a permittivity map. The GPRInvNet was able to make the best use of the GPR data and had accurately reconstructed the shapes and details of the defects. The high-quality dielectric maps were directly generated from the raw GPR data. Moreover, as we encoded the features trace by trace, the permittivity maps could be reconstructed column by column. Fig. 4 illustrates the schematic for two GPR traces in the GPRInvNet.

The architecture of the GPRInvNet, as well as its implementation, is described in the following section. Since the trace-to-trace encoder is specially designed for GPR data inversion, the encoding process was introduced in detail. However, the adopted decoder was conventional and topologically identical to the decoder component in the SeisInvNet model. Therefore, the decoder and the loss function of the network have only been briefly presented in this study.

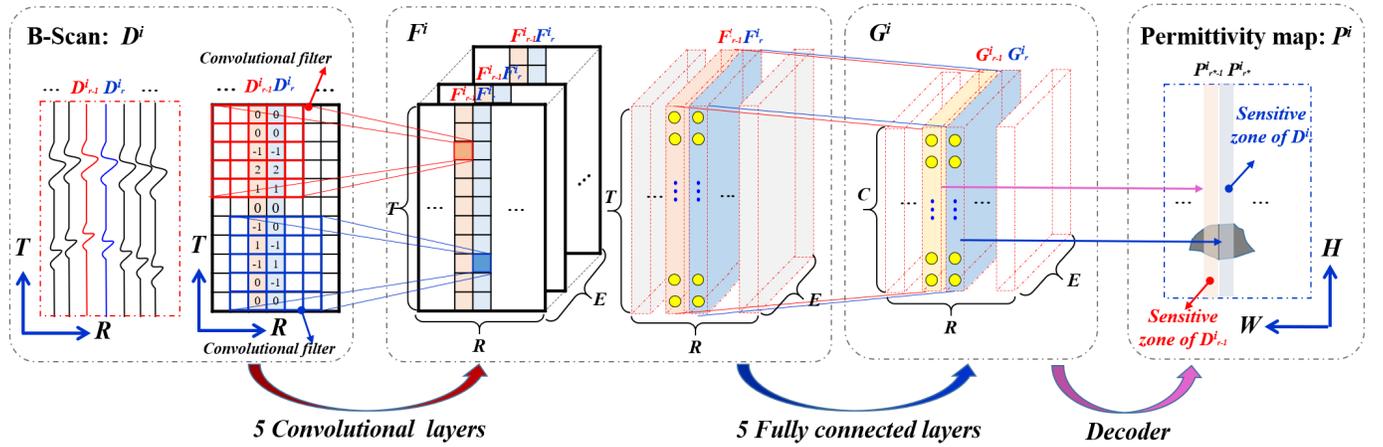

Fig. 4. Schematic for processing two GPR traces in the GPRInvNet. $D^i_{r-1}$ and $D^i_r$ are the two adjacent GPR traces of the B-Scan $D^i$; $F^i$ is the features extracted from $D^i$ by using five convolutional layers with $5 \times 5$ kernels; $F^i_{r-1}$ and $F^i_r$ are the $r-1$th and the $r$th columns of features of $F^i$. $F^i_{r-1}$ which is spatially aligned to $D^i_{r-1}$ is denoted as the "enhanced trace" of $D^i_{r-1}$, and $F^i_r$ which is spatially aligned to $D^i_r$ is denoted as the "enhanced trace" of $D^i_r$. $G^i$ is the features compressed from $F^i$ by using five fully connected layers. $G^i_{r-1}$ and $G^i_r$ are the two adjacent features of $G^i$. $G^i_{r-1}$ is obtained by compressing the time dimension (time steps) of $F^i_{r-1}$ from $T$ to $C$, and $G^i_r$ is obtained by compressing the time dimension (time steps) of $F^i_r$ from $T$ to $C$; $P^i_{r-1}$ and $P^i_{r*}$ are the adjacent columns in the permittivity map $P^i$. $P^i_{r-1}$ which is spatially aligned to $G^i_{r-1}$ is denoted as the "sensitive zone" of $D^i_{r-1}$, and $P^i_{r*}$ which is spatially aligned to $G^i_r$ is denoted as the "sensitive zone" of $D^i_r$.

## C. Trace-to-trace Encoder

In this study, the encoder of the GPRInvNet framework has been described as a trace-to-trace encoder. This is due to the fact that it was able to enrich the information of each trace and separately condense the features of each trace. The encoder consisted of five convolutional layers and five fully connected layers.

The convolutional layers were employed to generate the feature map $F$, which not only had the same dimensions as the input B-Scan images but also contained knowledge extracted from several adjacent GPR traces. This design was inspired by the fact that the echoes of one abnormality were often mainly observed in several adjacent traces of the GPR B-Scan. The numbers of convolutional layers of the network determined the numbers of adjacent traces from which we extracted the information related to the abnormality. However, the more convolutional layers we employed in the network, the more redundant parameters to be determined. Therefore, the GPRInvNet with different numbers of convolutional layers were compared. The encoder consisted of three, five, and seven convolutional layers was utilized to enrich the information of each GPR trace by extracting information from different numbers of adjacent GPR traces. The detailed framework of the encoder with different convolutional layers can be found in Fig. 5. Once the trace separation of the B-Scan was confirmed, the numbers of corresponding neighboring traces for three, five, and seven convolutional layers using $5 \times 5$ convolutional kernels were 13, 21, and 29, respectively. For the encoder with seven convolutional layers, we output the feature map with a dimension of $800 \times 99 \times 64$ so as to reduce the parameters in the network and save computational cost.

To illustrate what the "enriched GPR trace" looks like, we visualized the overall perspective for the feature map $F^i_{T \times R \times E}$ by computing the sum of feature values per pixel over all the $E$ channels ($E$ represents the numbers of feature channels of the feature map $F^i_{T \times R \times E}$). Fig. 5 depicts the comparative results of visualized features for different numbers of convolutional layers. As can be seen from Fig. 5, the feature map extracted from the B-Scan using the encoder with three convolutional layers retained most of the properties of the B-Scan. Compared with the groundtruth, "hyperbolic" morphologies can be observed around the zones where the abnormalities existed. While for the feature maps obtained by the encoder with five and seven convolutional layers, the hyperbolic morphologies were mitigated notably. The zones where the abnormalities existed have been highlighted, which means the "enhanced GPR traces" (extracted features spatially corresponding to the GPR traces) can provide more information to reconstruct the permittivity of the abnormalities. The feature maps extracted using seven convolutional layers were quite similar to those extracted using five convolutional layers, which mean there were no great differences between the information enriched using five convolutional layers and seven convolutional layers. Therefore, for the GPR data in this study, the information extracted from 21 neighboring traces was sufficient to enhance the information of each GPR trace. In order to save computational cost, encoder with five convolutional layers was employed.

The $5 \times 5$ convolutional kernels with a stride of 1 were employed in each layer in order to extract adjacent information from the GPR B-Scan $D^i_{T \times R \times 1}$. Also, this study experimented with $3 \times 3$ convolutional kernels. However, the results were found to underperform those obtained by employing $5 \times 5$ convolutional kernels. The receptive field for the $5 \times 5$ convolutional kernels with five convolutional layers was 21, while that for the $3 \times 3$ convolutional kernels was 11.

After five convolutional layers, a feature map $F^i_{T \times R \times E}$ with the same spatial dimensions as the input GPR B-Scans $D^i_{T \times R \times 1}$ was generated. In the current study, $F^i_{T \times R \times E}$ signifies the feature maps encoded from the *i*th GPR B-Scan, and $E$ denotes the number of feature channels. The obtained feature map $F$ had the same dimensions as the input. However, the feature at each position also contained the need neighborhood information. In regard to the GPR data of each trace $D^i r$ with the dimensions $[T; 1]$, the convolutional layers had converted the data into a feature vector $F^i r$ with the dimensions $[T; E]$. Then, each $F^i r$ was spatially aligned to a column of the permittivity map in order to complete the inversion process. The feature map $F^i_{T \times R \times E}$ was treated as the $R$ columns of the feature vectors of the $r$th GPR trace $F^i r$.

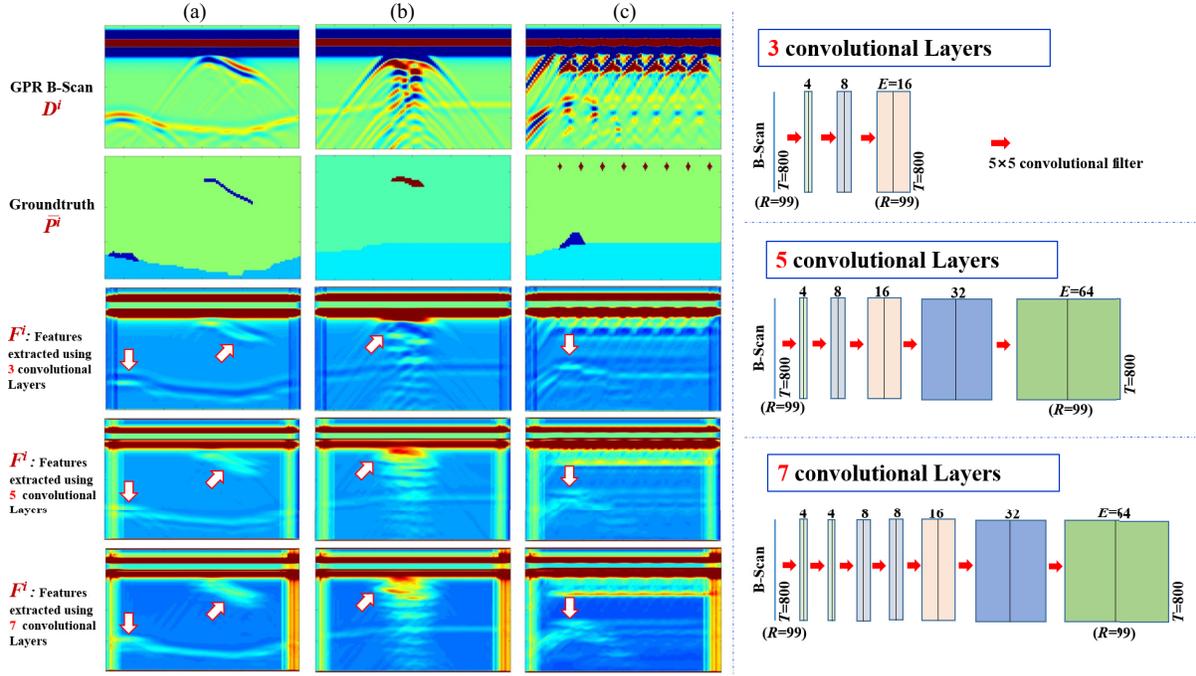

Fig. 5. Architecture of encoder with different convolutional layers and the corresponding visualized feature maps. The B-Scan $D^i$ with a dimension of $T \times R \times 1$ is the input; the output feature maps $F^i$ for encoder with three convolutional layers is with a dimension of $800 \times 99 \times 16$; $F^i$ for encoder with five convolutional layers is with a dimension of $800 \times 99 \times 64$; and $F^i$ for encoder with seven convolutional layers is with a dimension of $800 \times 99 \times 64$. (a) Visualized features for anhydrous defects. (b) Visualized features for water-bearing defects. (c) Visualized features for defects under rebar.

It was observed that, unlike the SeisInvNet model which encoded the embedded features into a feature map, the GPRInvNet encoded the features of each trace separately and then spliced the features of all the traces to form a group of feature maps. This process was implemented using the fully connected layers to separately condense the feature vector of each trace $F^i r$. The design of the fully connected layers in GPRInvNet originated from the need to maintain the spatial alignments between the input B-Scan images and output permittivity maps. In this study, for each encoded GPR trace $F^i r$ with the dimensions $[T; E]$, five fully connected layers were adopted to fuse the time dimensional features of each trace and combine them into features maps with the dimensions $[C; E]$. In addition, each fully connected layer included activation and batch normalization operations. The fully connected layers were implemented for all of the $R$ feature vectors of the feature map $F^i_{T \times R \times E}$. In this way, new feature maps $G^i_{C \times R \times E}$, which had the same dimension ratio as the permittivity map $P^i$, were generated. The $r$th column vectors of the new feature maps were denoted as $G^i r$, where

$G^i r$ indicates a size of $C \times E$. This study then artificially enforced each $G^i r$ to be spatially aligned to a column of the permittivity map to be inverted.

The trace-to-trace encoder adopted in this study had displayed the following two benefits for the GPR inversion in comparison to the existing networks: 1) the encoder had made the best use of the raw GPR data by using the convolutional layers to enhance the effective information of each trace from its adjacent traces and 2) it had retained the spatial alignments between the B-Scan images and permittivity map by separately encoding the features of each trace.

## D. Decoder and Loss Function

In the encoder, the feature maps $G^i_{C \times R \times E}$ with the same dimensional ratios as the permittivity maps to be inverted $P^i_{H \times W}$ are generated. More importantly, the features of each GPR trace $G^i r$ were spatially aligned to the sensitive zones in the permittivity map $P^i r^*$, where $r^* \in [I, W]$. Then, using those features, the high-quality dielectric images could be easily and accurately reconstructed.

The decoder, which was employed in this study, was similar to that of the SeisInvNet model [37]. However, adjustments made the parameters of the decoder networks based on the permittivity maps to be reconstructed. The decoder consisted of a $4 \times 4$ up-convolution, six $3 \times 3$ convolutions, and one up-sampling operation. A $4 \times 4$ up-convolution with a stride of 2 was first deployed in order to enlarge the dimensions of the feature maps. This was followed by a $3 \times 3$ convolution with a stride of 1, which was used to stabilize the information. Then, an up-sampling operation was utilized to form feature maps with the same dimensions as the permittivity maps. Finally, four $3 \times 3$ convolutional kernels with a stride of 1 were added for the purpose of condensing the dimensions of the feature channels. A dropout method was employed to randomly abandon some of the feature maps in order to avoid over-fitting and improve the robustness of the networks. In principle, each feature map $\boldsymbol{G^i r}$ with the size $C \times E$ was utilized to accurately invert a small piece of the permittivity model $\boldsymbol{P^i_{r^*}}$, which had the size $H \times 1$. The entire permittivity map $\boldsymbol{P^i}$ was accurately reconstructed in this study by splicing all of the $\boldsymbol{P^i_{r^*}}$ together.

In regard to the loss function, this study employed a combination of the L2 norm and multi-scale structural similarity (MSSIM) to minimize the misfits between the input and output images. The loss function was calculated as follows [37], [46]:

$$
\begin{aligned}
L_{\mathrm{i}}(P^i, \bar{P}^i) = {}& \sum_{h=1}^{H} \sum_{w=1}^{W} \left\| P^i_{(h,w)} - \bar{P}^i_{(h,w)} \right\|_2 \\
& - \sum_{r \in R} \sum_{h=1}^{H} \sum_{w=1}^{W} \lambda_r \cdot \mathrm{SSIM}\left( P^i_{x_{((h,w),r)}} - \bar{P}^i_{y_{((h,w),r)}} \right)
\end{aligned}
\tag{2}
$$

where $P^i$ and $\bar{P}^i$ represent the inversion results and ground truth for $i$th data pairs, respectively; $x_{((h,w),r)}$ and $y_{((h,w),r)}$ are the two corresponding windows centered on $(h, w)$ with size $r$, in which $h \in [1, H]$, $w \in [1, W]$; $R$ is the total number of scales; and $\lambda_r$ is the weight of scale $r$. Then, by simultaneously minimizing the norm metrics and maximizing the MSSIM, the model was optimized in both structural similarity and per pixel error rates in the output images.

## IV. NUMERICAL SIMULATION

### A. Building the Data Set

To provide sufficient data for the training of the GPRInvNet, numerical simulations were conducted to generate synthetic data for the common types of tunnel lining defects. These included lining voids, cracks, lining-rock delamination, leakages, and non-compactness of the concrete. The data set basically covered the most common types of tunnel lining defects characterized by irregular shapes. More specifically, the data set consisted of four categories with different configurations of defects as follows: 1) tunnel linings (concrete and rock) containing only rebars or one type of defect (lining voids, cracks, lining-rock delamination, or non-compactness); 2) tunnel linings with both rebar layers and one type of defect; 3) tunnel linings containing multiple defects; and 4) tunnel linings with rebar layers and multiple defects. For each category, both air and water were considered as the media of the defects, and the permittivity tendencies for the ground and concrete were chosen from a certain range. In total, a data set containing 432,000 pairs of data was built. Each data pair included a GPR B-Scan as the input and a permittivity model as the ground truth. There were five different types of media involved: air, surrounding rock, concrete, water, and rebars. The parameters utilized in the simulations are listed in Table I, which had been selected and modified from [47].

### TABLE I
RELATIVE DIELECTRIC CONSTANT AND CONDUCTIVITY PROPERTIES

| Media | Relative dielectric constant | Conductivity S/m |
|---|---|---|
| Air | 1 | 0 |
| Water | 81 | 0.0005 |
| Concrete | 8~10 | 0.0001 |
| Surrounding rock | 6~8 | 0.001 |
| Rebar | 300 | $10^{\wedge}8$ |

The GPR modeling was performed based on various finite difference time domain (FDTD) methods using an in-house MATLAB code. During the numerical simulations, a permittivity model with a width of 2.0 m and a depth of 0.7 m was constructed. A convolutional perfectly matched layer (CPML) was employed to mitigate the impacts of the boundary effects. The spatial size of the FDTD forward grid was 0.01 m, and the CPMLs had occupied 10 meshes. Therefore, each permittivity map consisted of $90 \times 220$ meshes, which included the $70 \times 200$ meshes of the tunnel lining and the outer 10 meshes of the CPMLs surrounding the tunnel lining. The total number of traces $R$ for each model is 99, and the total time step $T$ was 800, with a time window of $2.3587e^{-11}$s. The source wavelet was a Ricker wavelet with a center frequency of 600 MHz. The equation of the Ricker wavelet in time-domain can be expressed as follows:

$$
f(t) = (1 - 2\pi \cdot f_c^2 \cdot t^2) \cdot e^{-(\pi f_c \cdot t)^2}
\tag{3}
$$

where $f_c$ represents the center frequency.

One of the main contributions of this study was the reconstruction of the permittivity map of various tunnel lining defects with irregular geometries. Therefore, the shapes of the simulated lining defects were relatively irregular in order to show the applicability of the proposed GPRInvNet in the inversion of defects with various shapes. During the simulations, the background was first modeled, which included the concrete, surrounding rock, and rebars. The interfaces of the surrounding rock were generated by fitting randomly deployed nodes at the bottom of the model using the secondary spline curves. The rebar layers were located in the range of 5–25 cm, with intervals randomly selected between 15 and 30 cm. To simulated practical conditions of the tunnel, lining had resembled real situations as closely as possible. The dielectric parameters of the concrete and rock were randomly selected from the ranges listed in Table I.

Following the completion of the background modeling, the common types of tunnel lining defects with irregular shapes were generated by fitting the constraint nodes using the spline curves. The sizes of the voids had ranged from $16 \times 5$ cm to $60 \times 40$ cm, and the lengths of the cracks were between 20 and 60 cm. Then, in order to simulate the lining-rock delamination defects, this study first randomly located their positions along the interfaces between concrete and rock. The same method which had been adopted for the modeling process was implemented for the void defects in order to formulate delamination defects with sizes ranging

| Stage | Layer | Type | Filter | Stride | Output Size |
|-------|-------|------|--------|--------|-------------|
| Input | - | - | - | - | 800×99×1 ($\boldsymbol{D}^i:T{\times}R{\times}I$) |
| Encoder | L1 | conv | 5×5 | 1 | 800×99×4 |
| | L2 | conv | 5×5 | 1 | 800×99×8 |
| | L3 | conv | 5×5 | 1 | 800×99×16 |
| | L4 | conv | 5×5 | 1 | 800×99×32 |
| | L5 | conv | 5×5 | 1 | 800×99×64 ($\boldsymbol{F}^i:T{\times}R{\times}E$) |
| | L6 | FC | 1024 | 1 | 1024×99×64 |
| | L7 | FC | 512 | 1 | 512×99×64 |
| | L8 | FC | 256 | 1 | 256×99×64 |
| | L9 | FC | 256 | 1 | 256×99×64 |
| | L10 | FC | 45 | 1 | 45×99×64 ($\boldsymbol{G}^i:C{\times}R{\times}E$) |
| Decoder | L11 | Up-conv | 4×4 | 2 | 90×198×128 |
| | | conv | 3×3 | 1 | 90×198×128 |
| | L12 | Upsample | - | - | 90×220×128 |
| | | conv | 3×3 | 1 | 90×220×64 |
| | L13 | conv | 3×3 | 1 | 90×220×64 |
| | | conv | 3×3 | 1 | 90×220×32 |
| | L14 | conv | 3×3 | 1 | 90×220×32 |
| | | conv | 3×3 | 1 | 90×220×1 |
| Output | - | - | - | - | 90×220×1 ($\boldsymbol{P}^i:H{\times}W{\times}I$) |

from $16 \times 5$ cm to $100 \times 40$ cm. In regard to the noncompactness defects, sections with sizes ranging from $20 \times 20$ cm to $60 \times 60$ cm were randomly selected. Subsequently, many small voids were generated within the selected sections. All of the defects were randomly distributed in the permittivity map.

### B. Experimental Process

This study randomly assigned each configuration of defects to the training data, validation data, and testing data with a set ratio of 10:1:1. Then, a data set with a total of 432 000 data pairs was randomly divided into three sub-data sets, in which 360 000 pairs were used for training, 36 000 for validation, and 36 000 for testing. In order to verify the impact of the conductivity on the inversion result of relative permittivity, a supplementary data set with a total of 50 data pairs was added into the testing data set. The defects of the permittivity models in the supplementary data set were filled with a media with the conductivity of 0.5 S/m and the relative dielectrics constant of 81. It should be noted that the permittivity models in the supplementary data set contained different electric conductivities, and the GPRInvNet was never trained on the supplementary data set. Each B-Scan $\boldsymbol{D}^i$ was measured $800 \times 99$ ($T \times R$), and the corresponding permittivity model $P^i$ had the dimensions of $90 \times 220$ ($H \times W$). The outer ten layers of $\boldsymbol{P}^i$ were CPMLs, which had been cropped from the permittivity model. Finally, an output permittivity model with the dimensions $70 \times 200$ was obtained, which corresponded to a tunnel lining measuring $0.7 \times 2$ m.

The details of each layer of the GPRInvNet in this study are listed in Table II. The experiments were conducted on an Intel

Xeon (R) Gold 5118 CPU workstation, with 64 GB RAM and a GTX 1080 Ti GPU. The GPRInvNet was implemented based on Pytorch [48]. Then, in order to optimize the GPRInvNet, an Adam optimizer with batch size of 12 was applied, with a learning rate of $5e^{-5}$. The dropout rate in the decoder was 0.2. The GPRInvNet contained 2 041 326 parameters and could be trained end-to-end. The models were trained for 100 epochs (means 100 iterations in this study), which was observed to be more than sufficient to ensure convergence. In addition, following the convention, the parameters which had performed best on the validation set were saved and follow-up experiments were conducted on the validation and test sets.

A series of metrics were employed to quantitatively evaluate the performance results of the GPRInvNet. This study quantified the misfit errors of the inversion results based on mean average error (MAE) and mean square error (MSE) methodology [37]. In addition, the similarities of the local structures were measured by MSSIM [45] and SSIM [49].

### C. Comparative Study and Results

For the purpose of verifying the superiority of the proposed method in reconstructing the permittivity maps of tunnel linings containing complex internal defects, a comparative study was performed based on synthetic GPR data. Both physics-driven and data-driven methods were chosen as baselines. The GPRInvNet was compared against the DNN-based model Enc-Dec, as well as the widely used physics-driven method FWI. All of the methods were tested on the same testing data set. This study's comparative results are shown in Figs. 6 and 7. The inversion results for the relatively simple tunnel lining defects are displayed in Fig. 6(a-1)–(e-1), and Fig. 7(a-1)–(e-1) illustrates the ground truths. The permittivity maps reconstructed using the FWI are shown in Fig. 6(a-2)–(e-2) and Fig. 7(a-2)–(e-2). The inversion results of the Enc-Dec are detailed in Fig. 6(a-3)–(e-3) and Fig. 7(a-3)–(e-3). In addition, Fig. 6(a-4)–(e-4) and Fig. 7(a-4)–(e-4) provide the reconstruction results achieved using the GPRInvNet.

### D. Comparison With Enc-Dec Network

The Enc-Dec network, which was initially developed for image segmentation, was employed for mapping the GPR B-Scan images to the subsurface images of concrete in [42]. The image-to-image translations were found to be similar to the inversion task, with the exception that the inversion provided not only the subsurface images but also the permittivity values. The Enc-Dec network was found to outperform the U-Net and GAN for imaging the subsurface defects according to the aforementioned study. Therefore, the Enc-Dec network was chosen as a baseline model in this article.

The Enc-Dec in this study had the same architecture as in [35] and [36]. It consisted of an encoder and a corresponding decoder. Each convolutional layer employed a $4 \times 4$ convolutional filter with a stride of 2 for downsampling the input images, batch normalization, and element-wise rectified-linear nonlinearity (ReLU) in order to extract features from the GPR B-Scan images. Accordingly, the decoder recovered the defects from the embedding vector. Then, $4 \times 4$ transposed convolutions with a stride of 2 were employed to upsample the vector. The loss function was DSSIM [42], and the initial learning rate was set as $5e^{-5}$.

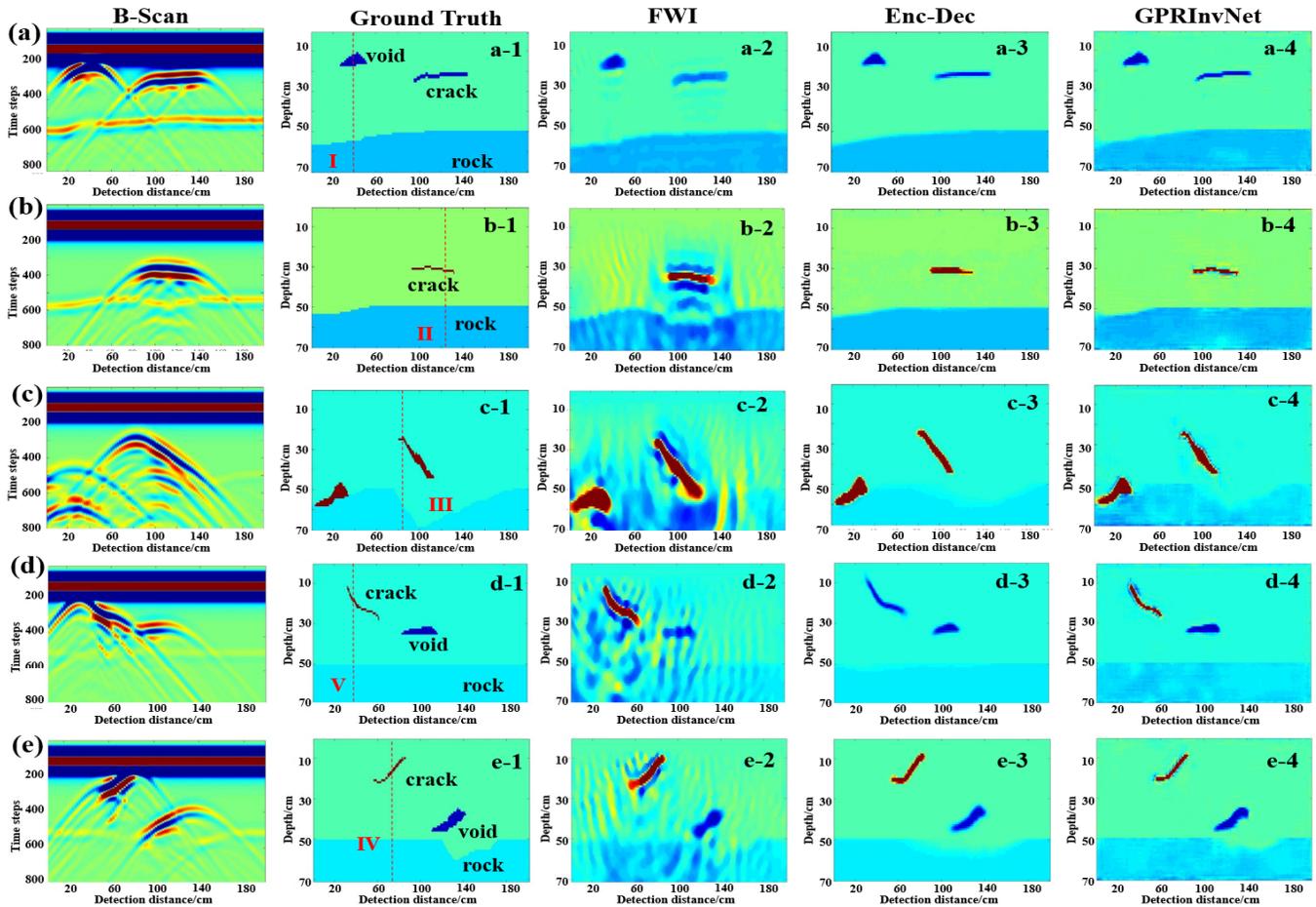

Fig. 6. Inversion results for simple tunnel lining defects. (a) Tunnel lining with anhydrous defects. (b) Tunnel lining containing one water-bearing defect. (c) Tunnel lining with multiple water-bearing defects. (d) Tunnel lining containing multiple defects with different relative permittivity and different conductivities. (e) Tunnel lining containing multiple defects with different relative permittivity and different conductivities. The images of the first column are B-Scan; the images of the second column are ground truths; the inversion results of the FWI, Enc-Dec, and GPRInvNet are illustrated in the images of the third, fourth, and fifth columns, respectively; and lines I–V are the four cutting lines.

Both the Enc-Dec and GPRInvNet were trained on the same data set for 100 epochs. In addition, in order to maintain physical information in the input and output data, this study directly employed the raw data pairs to train the network rather than reformatting them into images. As a result, the input and output of the Enc-Dec network in this study both contained data of the physical characteristics of the concrete rather than the pixels of the images. The input B-Scan data with the dimensions of $T \times R$ were resized into a matrix with dimensions of $256 \times 128$, and the output permittivity maps were resized from [128, 256] to [90, 220].

As can be seen from Fig. 6, Enc-Dec was capable of reconstructing the interfaces between rock and concrete as well as some simple defects, such as cracks, voids, and delamination without rebar. However, the reconstructed defects typically had blurred boundaries. Moreover, for some of the permittivity maps containing multiple defects with different relative permittivity and different conductivity, Enc-Dec incorrectly reconstructed the water-bearing defect, as shown in Fig. 6(d-3). In contrast, GPRInvNet successfully reconstructed tunnel lining defects with relatively clear boundaries. The shapes of anhydrous cracks, anhydrous voids, water-bearing cracks, rebar, and surrounding rocks reconstructed by GPRInvNet were highly consistent with the ground truths. Even when the materials of the multiple defects inside the permittivity model are different (one defect was filled with water and the other was air) and present noticeably different electric conductivities, the GPRInvNet could still reconstruct these tunnel lining defects with relatively clear boundaries [Fig. 6(d-4) and (e-4)]. Additionally, it should be noted that the permittivity models with new electric conductivities were directly utilized for testing, and the GPRInvNet was never trained on the data pairs with new electric conductivities. The result indicates that the GPRInvNet has satisfying adaptability.

For tunnel linings with complex defects, such as noncompactness, defects under the rebar, etc., GPRInvNet clearly outperformed Enc-Dec. As can be seen in Fig. 7(b)–(d),

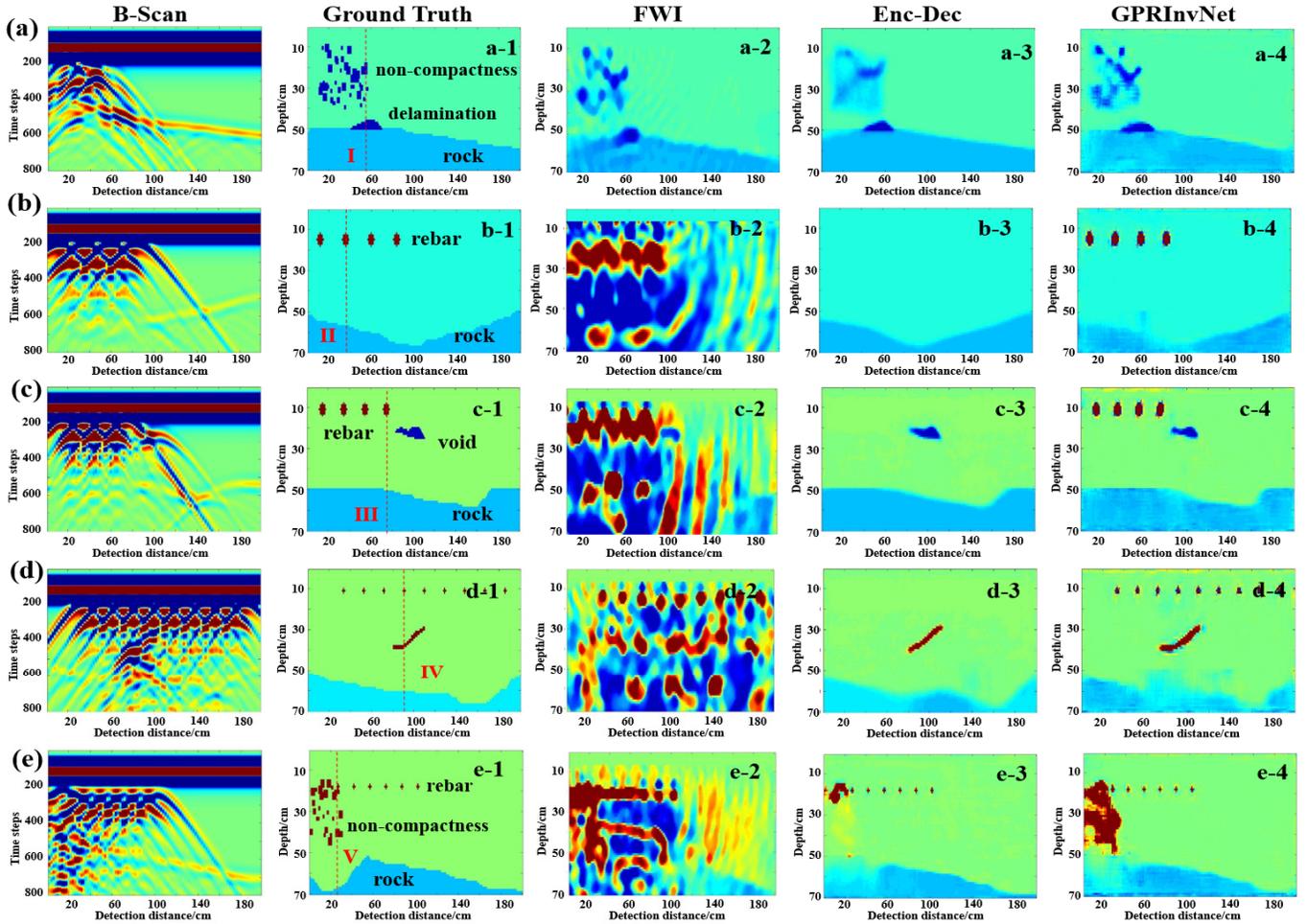

Fig. 7. Inversion results for complex tunnel lining defects. (a) Tunnel lining with noncompactness and delamination. (b) Tunnel lining with rebar only. (c) Tunnel lining with anhydrous defects under rebars. (d) Tunnel lining containing one water-bearing crack under rebars. (e) Tunnel lining with water-bearing noncompactness under rebars. The images of the first column are B-Scan; the images of the second column are ground truths; the inversion results of the FWI, Enc-Dec, and GPRInvNet are illustrated in the images of the third, fourth, and fifth columns, respectively; Lines I–V are the four cutting lines.

Enc-Dec failed to recover the rebar for most of the cases. This is probably because the size of each rebar is very small, so Enc-Dec failed to learn the features of rebar during the feature extraction process. The Enc-Dec reconstructed the rebar in Fig. 7(e-3). This is probably because the size differences of the water-bearing non-compactness defects and the rebar were small in this case, and their characteristics were quite similar. The Enc-Dec learned effective features for this particular case during the training process. For the noncompactness [Fig. 7(a)] and water-bearing non-compactness defects under rebar [Fig. 7(e)], because the honeycomb obstructed the propagation of electromagnetic waves, neither method could perfectly recover the shapes of defects. As shown in Fig. 7(a) and (e), Enc-Dec could only partially recover the defect with a very blurry region. In contrast, GPRInvNet successfully provided complete profiles of the complex non-compactness defects [Fig. 7(a-4)] even if the defects are below the rebar [Fig. 7(e-4)]. Although the shapes of the complex defects reconstructed by GPRInvNet were not completely consistent with the ground truths, it still provided the best inversion results.

TABLE III
COMPARISON OF EVALUATION METRICS

| Dataset | Metrics | GPRInvNet | Enc_Dec |
|---------|---------|-----------|---------|
| Valid | MAE ↓ | **0.002845** | 0.004927 |
| | MSE↓ | **0.000368** | 0.002539 |
| | SSIM↑ | **0.973808** | 0.949326 |
| | MSSIM↑ | **0.980597** | 0.855101 |
| Test | MAE ↓ | **0.002860** | 0.004895 |
| | MSE↓ | **0.000374** | 0.002515 |
| | SSIM↑ | **0.973784** | 0.949639 |
| | MSSIM↑ | **0.980623** | 0.858237 |

To quantitatively evaluate the performances of the two DNN-based methods, the evaluation metrics on the validation set and test set are listed in Table III. In general, GPRInvNet achieved the best performance. On the test set, the MAE, MES, SSIM, and MSSIM of GPRInvNet were 0.00286, 0.000374, 0.973784, and 0.980623, respectively. They are better than

Enc-Dec's, which were 0.004895, 0.002515, 0.949639, and 0.858237. In general, GPRInvN showed consistent superiority according to all evaluation metrics compared.

Therefore, GPRInvNet outperformed Enc-Dec in GPR inversion. The shapes and details of the defects reconstructed by GPRInvNet are obviously better than those reconstructed by Enc-Dec. This is due to the specially designed encoding approach of GPRInvNet, which can make full use of the information during the extraction of the feature map as well as retaining the spatial alignment between the input and output. On the contrary, Enc-Dec employs a fixed convolutional kernel to extract the feature map as well as compressing the dimension. This may lose detailed information, such as the features of small-sized rebars and the boundaries of defects. Moreover, Enc-Dec decodes the permittivity map from the vector, which may contribute to the loss of spatial information.

### E. Comparison With FWI

FWI is used to reconstruct dielectric properties with all received waveforms, by means of minimizing the difference between the forward modeling waveform and observed waveform [13]. FWI for GPR data is very sensitive to the initial model, and is easily convergent to a local minimum or cycle skipping. Many developments of FWI have been presented, so as to improve its performance. In tunnel lining-related applications, an FWI with multiscale strategy was introduced in [24] to accurately reconstruct the complex irregular defects. In order to demonstrate the performance of GPRInvNet, in this section of this article, a comparative study against the multi-scale FWI method was performed. The multi-scale inversion strategy employed frequency information ranging from low-to high-frequency regions during the inversion process, so as to avoid FWI convergence to a local minimum or cycle skipping. In addition, in order to further improve the performance, the initial models utilized in this study were obtained by applying a Gaussian smoothing to the true permittivity models [50].

The multiscale FWI procedure in this study is as follows.

*Step 1:* Input the initial models. In this article, the initial models were obtained by applying a Gaussian smoothing to the true models. This operation has previously been utilized in [50] to improve the quality of the FWI of GPR data. The size of the square matrix in the Gaussian function was 35, and the standard deviation was set to 35. By doing so, the main trends of the defects and interfaces were retained in the initial models.

*Step 2:* Determine the number of multiple scales and the objective frequency of each scale. The multiscale inversion strategy decomposes the inversion problem into multiple scales with various frequencies. In this study, the inversion was performed by using the following three scales: 200 MHz for inversing the envelopes and 200 and 600 MHz for inverting the permittivity maps. The flowchart of the three scales is designed as follows.

1) The objective frequency was assigned as 200 MHz. Then, Steps 3–6 were employed to invert the envelopes using the low-frequency GPR data with the objective frequency of 200 MHz.

2) The inversion result of 1) was employed as the initial model, and the objective frequency was assigned as 200 MHz. Then, we repeated Steps 3–6 to invert the permittivity map using the low-frequency GPR data with the objective frequency of 200 MHz.

3) The objective frequency was assigned as 600 MHz and the inversion result of 2) was employed as the initial model. Next, the high-frequency GPR data with the objective frequency of 600 MHz were employed to invert the permittivity maps by repeating Steps 3–6 once again.

*Step 3:* Filter the GPR data using a Wiener low-pass filter. The multiscale inversion strategy involved a Wiener low-pass filter by which to process the wavelet of source pulse and GPR data, so as to obtain the information within the objective frequency. More specifically, the collected GPR data were transformed into the frequency domain. Subsequently, the Wiener low-pass filtering with the preset objective frequency in Step 2 was applied on the source pulse and the collected GPR data. The equation of Wiener low-pass filter can be found in [24]. Finally, the filtered results were transformed back into the time domain.

*Step 4:* Calculate the gradient using the equations of forward- and back-propagated wave-fields. Then, calculate the iteration stride based on the equation of gradient direction and update the model. The equations utilized in this step can be found in our previous study [25].

*Step 5:* Determine whether or not the termination condition has been achieved. The termination condition was set to achieve the maximum number of iterations at 200 times. If the termination condition was achieved, then proceed to Step 6, otherwise return to Step 4.

*Step 6:* Access if FWI has been conducted for all the scales. If not, return to Step 2 and update the initial model and objective frequency. If the FWI for multiple scales has been completed, then output the inversion results.

At this point, the multiscale FWI of GPR data has been realized using low- to high-frequency serial inversion techniques. The frequency-by-frequency strategy was capable of avoiding the local minimum, thereby retaining the details of the defect and eliminating the pseudomorph of inversion. Therefore, the FWI is shown to be suitable for the inversion of complex tunnel lining irregular defects.

The inversion results of FWI for tunnel lining with simple defects can be found in Fig. 6. As we can see from Fig. 6, FWI can roughly determine the profile of cracks, voids, and delamination in non-reinforced concrete with relatively blurred boundaries. The inversion results of the anhydrous defects [Fig. 6(a-2)] are slightly better than those of the water-bearing defects [Fig. 6(b-2) and (c-2)]. For the multiple defects filled with different materials, the FWI could reconstruct their profiles [Fig. 6(d-2) and (e-2)], although the boundaries are blurred. On the contrary, GPRInvNet could reconstruct the defects with clear boundaries.

For the complex defects, FWI provided the best inversion results in reconstructing the anhydrous noncompactness. It is outperformed the Enc-Dec, but it still underperformed GPRInvNet. Comparing Fig. 7(a-2) with Fig. 7(a-4), both the FWI and the GPRInvNet provide relatively accurate detailed structures of noncompactness. However, the FWI presents blurred profiles for the delamination under noncompactness. For defects under rebar, it is apparent that FWI underperforms the GPRInvNet for reconstructing anhydrous defects and water-bearing defects deployed in reinforced concrete, as is shown in Fig. 7(b-2), (c-2), (d-2), and (e-2). The FWI could only reconstruct the region of rebar. The defects under rebar were submerged in the pseudomorph, as shown in Fig. 7(c-2), and (e-2). This is probably due to the mask of rebar on the defects. Moreover, as there are great differences in the dielectric properties of the rebar and the defects, so most electromagnetic waves were reflected by rebar. In such situations, FWI could rarely achieve optimal parameters. However, GPRInvNet provided more satisfying results and clear boundaries of defects, even if the defects were under the rebar. Thus, the overall performance of GPRInvNet is better than that of the traditional FWI.

In terms of the computational cost, the multi-scale FWI took approximately 60 min to complete inversion for a single B-Scan image. On the contrary, a well-trained GPRInvNet is capable of inverting one image within approximately 0.027 s. Although training the neural network is a computationally intensive process, it takes place only once. The well-trained GPRInvNet can be used with near-real-time speed for GPR inversion.

### F. Comparison of Permittivity Values

To further analyze the inversion effects, we compared the permittivity values of the aforementioned methods. The relative permittivity values were extracted along the cutting lines. The cutting lines were numbered from I to V, as shown in Figs. 6 and 7. The relative permittivity values along the cutting lines of Figs. 6 and 7 are shown in Figs. 8 and 9 respectively.

As shown in Figs. 8 and 9, the inversion curves of permittivity through GPRInvNet are essentially the same as those of the ground truth, except that the amplitude is slightly different. Compared with Enc-Dec and FWI, we can find that for all the common defects, the oscillation of permittivity reconstructed by GPRInvNet is effectively alleviated and closer to the real values. Taking the line V in Fig. 9 as an example, GPRInvNet successfully predicted the variation of permittivity at a depth of 40 cm, while the remaining methods mis-detected it. FWI gave the worst results, as shown in Figs. 8 and 9 (pink line). It can be observed that the FWI provides desirable inversion results for the anhydrous defects. However, the relative permittivity values of defects under rebar and water-bearing defects predicted by FWI are always away from the true values. Compared with the DNN-based method, the permittivity curves reconstructed using traditional FWI show great fluctuations. The performance of Enc-Dec was better than that of FWI, but it still underperformed GPRInvNet. For some water-bearing defects, the results of Enc-Dec differed greatly from the real results in terms of the general trend of permittivity change (Fig. 9; Line V). Moreover, Enc-Dec failed to reconstruct some of the permittivity of rebar. Therefore, the overall performance of GPRInvNet was better than that of Enc-Dec as well as the widely used FWI method.

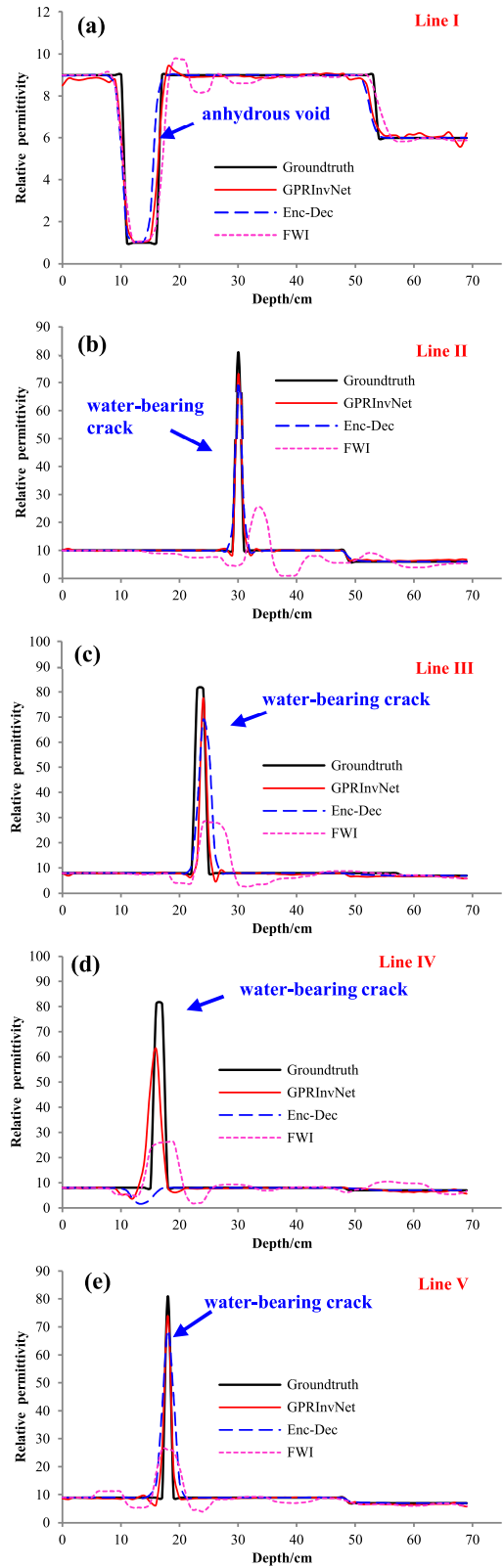

Fig. 8. Comparison of inverted permittivity values along the depth of the tunnel linings for simple tunnel lining defect. (a)–(e) Inverted relative permittivity values of cutting lines I–V in Fig. 6, respectively.

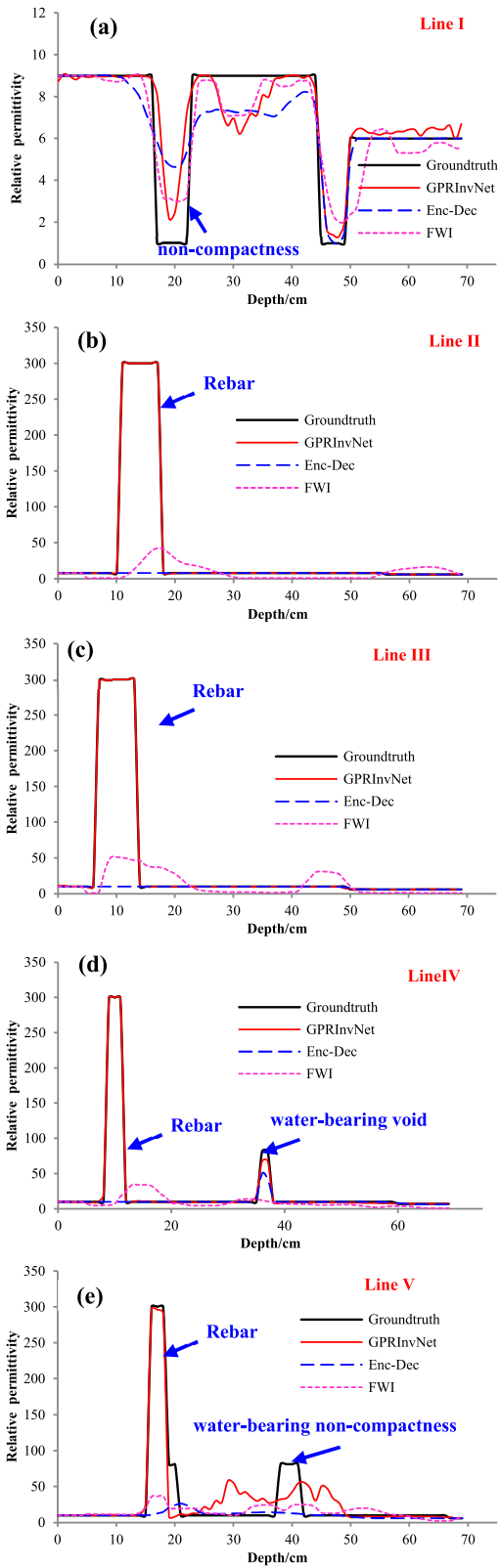

anhydrous defects lie in Fig. 9 Line I at a depth of 20 cm. The relative permittivity of the anhydrous defects inverted by GPRInvNet was approximately 2, which is close to the true value of 1, while at the same location, the inverted permittivity values by Enc-Dec and FWI were 4 and 3, respectively.

For GPRInvNet, the inverted value of rebar with the maximum deviation can be found in Line V of Fig. 9. The inverted value is 296, which is quite close to the true value of 300 in the simulation. On the contrary, the result of rebar provided by FWI with the best performance was 50, and Enc-Dec failed in recognizing rebar. For the water-bearing defects, the results reconstructed using GPRInvNet were also close to the true values. For example, the true relative permittivity at a depth of 24 cm in Fig. 8 Line III is 80. The permittivity provided by GPRInvNet was approximately 78, while the result of Enc-Dec was 68. Even for the water-bearing defects below the rebar (Fig. 9(e) at the depth of 40 cm), only GPRInvNet predicted the variation of the permittivity. Although the predicted value of 40 was smaller than the true value, it still performed better than other methods. In general, we can conclude that GPRInvNet outperforms other methods in quantitatively inverting permittivity values.

### G. Comparative Study of Resolution

The resolution of the GPR image as proposed in [51], can be understood as the capacity to discriminate individual targets in the subsurface. The vertical resolution is the capacity to discriminate two adjacent targets separated in different depths as two different events, while the horizontal resolution is the minimum distance between two targets located at the same depths that can be detected as two events [52], [53]. Unlike the geophysics model-driven inversion methods, it is difficult for the DNN-based inversion method which is data-driven to provide a mathematical relationship for imaging resolution. At present, there are no existing literatures to study the imaging resolution for the deep learning-based GPR inversion and imaging method, even in the field of seismic inversion and electrical resistivity inversion. In addition, in the studies related to deep learning-based images superresolution, which has been one of the most active research areas, the mathematical relationships have also not been derived and resolutions of all the methods were validated on the classical data sets [54], [55]. Therefore, in order to verify the resolution of the proposed method, we examined the performance of the classical square models and rebar models. These models have been documented and were considered as effective indicators of resolution issues [56], [57]. A supplementary data set with square models and rebar models was synthesized to fine-tune the GPRInvNet. The supplementary data set consisted of 15 600 data pairs, in which 11 388 for training, 2106 for validation, and another 2106 for testing. The bandwidth and central frequency of the GPR excitation signal are important for characterizing the resolution [59], [61]. In order to study the effect of different GPR excitation signals on the resolution, three different GPR source wavelets, which were widely utilized for tunnel detection, were employed to synthesize the GPR data in this study. The three source wavelets were the Ricker wavelets [60] with central frequencies of 400, 600,

Fig. 9. Comparison of inverted permittivity values along the depth of the tunnel linings for complex tunnel lining defects. (a)–(e) Inverted relative permittivity values of cutting lines I–V in Fig. 7 respectively.

We also looked closer at the permittivity measured for different filling materials and compared the inversion results at a pixel level. The inverted results with maximum error for and 900 MHz. The bandwidths were calculated by referring to [61], and the corresponding bandwidths for the 400, 600, and 900 MHz Ricker wavelets were approximate 436, 654, and 981 MHz, respectively. The time-domain signals and amplitude spectrums of Ricker wavelets employed in this study are illustrated in Fig. 10(a) and (b). The spatial size of the FDTD forward grid for the models of the three central frequencies was 0.01 m, and the dimensions of the permittivity map of all of the models were 70 cm × 200 cm. Within the models, two targets (two square abnormalities or two rebars) with different distances and depths were randomly deployed into the models [56], [57]. In order to clearly illustrate the targets, we zoomed in on the models and concentrated on the region of 70 cm × 70 cm in the middle of the model, as shown in Figs. 11 and 12. Both air and water were successively filled in the square abnormalities as anhydrous and water-bearing defects. The relative permittivities of the air, water, rebar, and background were 1, 81, 300, and 9, respectively. The dimensions of the square abnormalities were randomly selected from 10 cm × 10 cm–25 cm × 25 cm, and the minimum diameters of rebars were 1 cm. The parameters of GPRInvNet were fine-tuned on the supplementary data set for 50 epochs, and the batch size for each epoch was 12.

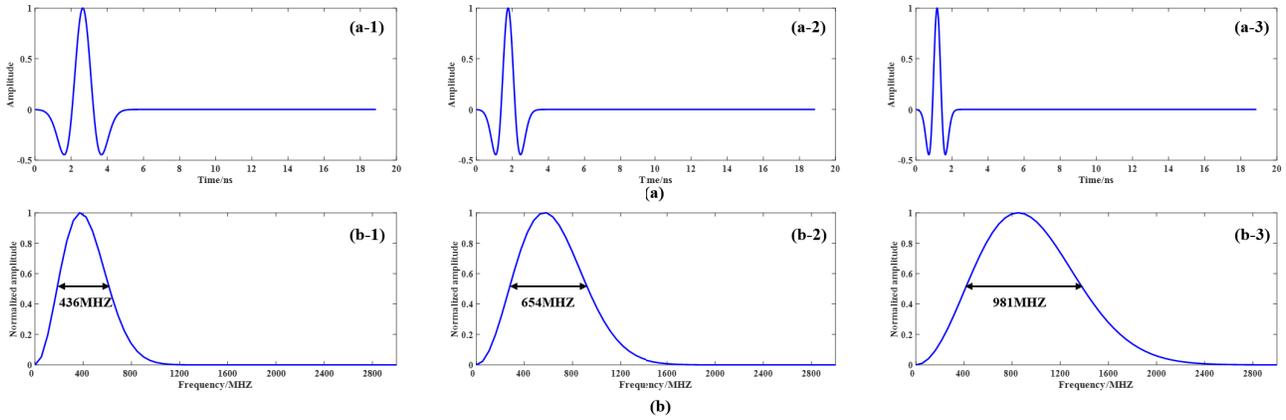

Fig. 10. Time-domain signals and the amplitude spectrums of Ricker wavelets. (a) Time-domain signals of Ricker wavelets with different central frequencies. (b) Normalized amplitude spectrums of Ricker wavelets with different central frequencies. (a-1)–(a-3) Time-domain signals of Ricker wavelets with central frequencies of 400, 600, and 900 MHz, respectively. (b-1)–(b-3) Normalized amplitude spectrums of the Ricker wavelets with central frequencies of 400, 600, and 900 MHz, the corresponding bandwidths for the 400, 600, and 900 MHz Ricker wavelets were approximate 436, 654, and 981 MHz, respectively.

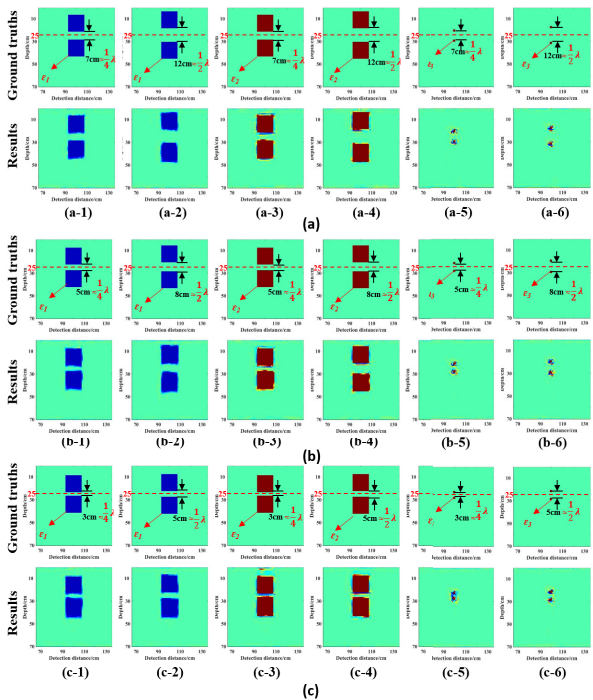

Fig. 11. Inversion results of square models and rebar models with different bandwidths and central frequencies to verify the vertical resolution. The dimensions of the square models are 15 cm × 15 cm, and the diameters of the rebars are 2 cm. The red horizontal lines represent the medial depths of targets, which are 25 cm. The first and second columns are anhydrous abnormalities. The third and fourth columns are water-bearing abnormalities. The fifth and sixth columns are rebars. (a) Ground truths and the corresponding inversion results of the 400-MHz GPR data. (b) Ground truths and the corresponding inversion results of the 600-MHz GPR data. (c) Ground truths and the corresponding inversion results of the 900-MHz GPR data. $\varepsilon_1$, $\varepsilon_2$, and $\varepsilon_3$ are the relative permittivities, where $\varepsilon_1$ is 1, $\varepsilon_2$ is 81, and $\varepsilon_3$ is 300.

We gradually increased the distances between the two targets, and some inversion results for the cases with different vertical resolutions were illustrated in Fig. 11. It can be seen in Fig. 11, for the GPR data with different central frequencies (corresponding to different bandwidths), the GPRInvNet was capable of distinguishing anhydrous targets, water-bearing targets, and rebars deployed with the vertical interval of 7, 5 and 3 cm, respectively [Fig. 11(a-1), (a-3), (a-5), Fig. 11(b-1), (b-3), (b-5), and Fig. 11(c-1), (c-3), (c-5)]. When we increased the distances, the boundaries of the two targets can also be detected clearly. Although the inverted shape was slightly blurred for the rebars, the two targets can still be detected. The results indicate that the GPRInvNet can effectively reconstruct the targets filled with air, water, and metal, and the detectable vertical distances for the two targets were slightly larger than a quarter of the wavelength.

Fig. 12(a)–(c) illustrate the inversion results for GPR data with different central frequencies and bandwidths, in which two targets were placed horizontally with different intervals to verify the horizontal resolution. The results show that GPRInvNet can distinguish anomalies with depths ranging

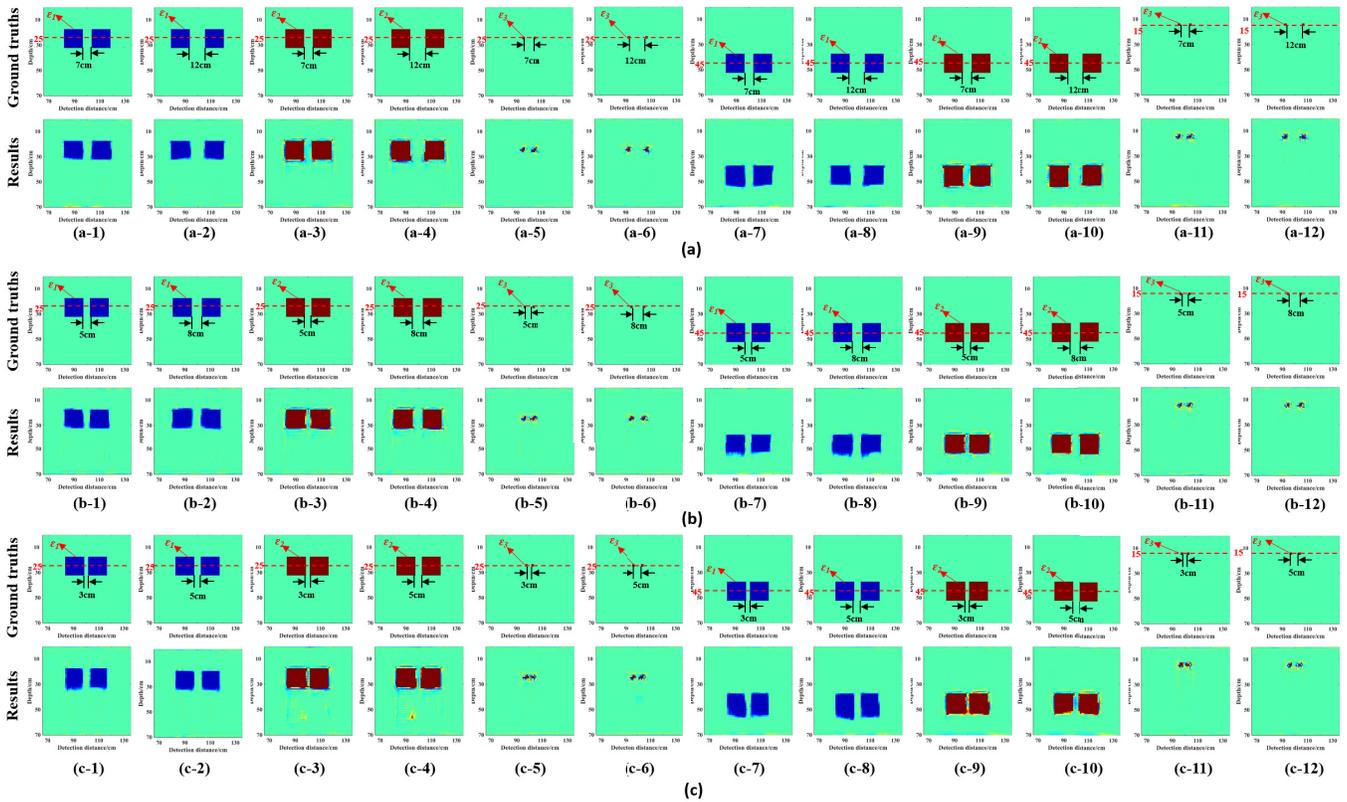

Fig. 12. Inversion results of square models and rebar models with different bandwidths and central frequencies to verify the horizontal resolution. The dimensions of the square models are 15 cm × 15 cm, and the diameters of the rebars are 2 cm. The red horizontal lines represent the medial depths of targets. The depths of 1st–6th columns are 25 cm, the 7th–12th columns are 45cm, and the 11th–12th columns are 15 cm. The first and second columns are anhydrous abnormalities. The third and fourth columns are water-bearing abnormalities. The fifth and sixth columns are rebars. (a) Ground truths and the corresponding inversion results of the 400-MHz GPR data. (b) Ground truths and the corresponding inversion results of the 600-MHz GPR data. (c) Ground truths and the corresponding inversion results of the 900-MHz GPR data. $\varepsilon 1$, $\varepsilon 2$, and $\varepsilon 3$ are the relative permittivities, where $\varepsilon 1$ is 1, $\varepsilon 2$ is 81, and $\varepsilon 3$ is 300.

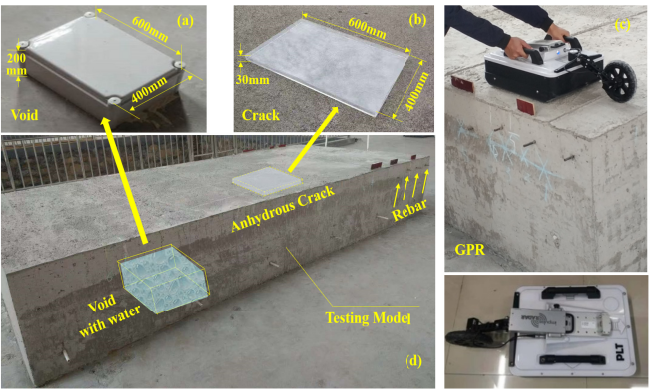

Fig. 13. Schematic of testing model and defects. (a) Waterproof box for the simulating void. (b) Acrylic box for simulating the cracks. (c) GPR system utilized in the experiment. (d) Model testing results and the deployment of the defects.

from 15 to 45 cm. As it can be seen in Fig. 12(c-7)–(c-10), the GPRInvNet is capable of detecting targets at the depths of 45 cm with an interval of 3 cm [Fig. 12(c-7) and (c-9)] and 5 cm [Fig. 12(c-8) and (c-10)]. However, the shapes of the reconstructed targets at the depths of 45 cm were slightly blurred than at the depths of 25 cm [Fig. 12(c-1)–(c-4)].

A similar conclusion can be found for the GPR data with the central frequencies of 400 and 600 MHz. It demonstrated that horizontal resolutions are related to the depths, and the shallow targets have higher horizontal resolutions, which agrees with the theoretical analysis [56].

The results indicate that for the inversion of tunnel lining, the GPRInvNet can take full use of the high-frequency information from the full waveforms of GPR data [17], [58], and provide desirable resolution (slightly larger than a quarter of the wavelength). Moreover, it was also found that when the vertical or horizontal distance between the two targets is less than a quarter of the wavelength, that GPRInvNet cannot clearly distinguish the two targets.

## V. EXPERIMENTS ON REAL DATA

The previous section demonstrated the superiority of the GPRInvNet when synthetic data had been used. However, the question remained as to whether the GPRInvNet was applicable to real data. It has been found that real GPR data is much more complex, and currently, there is no available real data for training data-driven DNN models. In this section, a method was presented for generalizing the GPRInvNet trained using synthetic data on real data. An experimental modeling process was performed to validate the feasibility of the GPRInvNet in such cases.

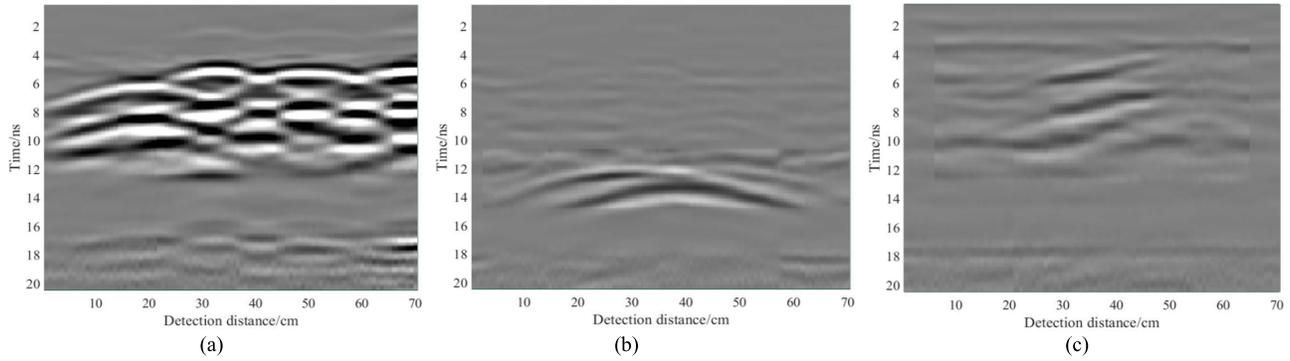

Fig. 14. GPR B-Scan images from the experiments. (a) B-Scan images of the rebars. (b) B-Scan images of the water-bearing defect. (c) B-Scan images of the anhydrous defect.

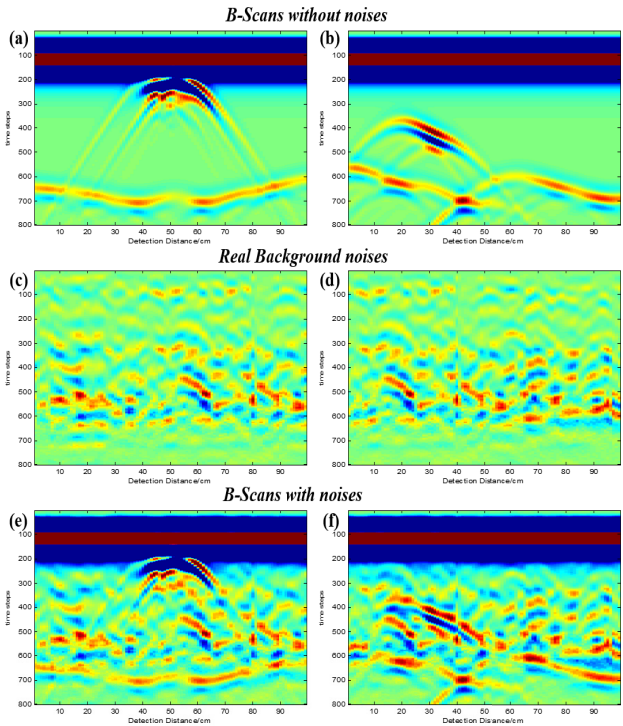

Fig. 15. B-Scan data with background noise patches. (a) and (b) Two synthetic B-Scan data. (c) and (d) Two real background noises. (e) Synthetic B-Scan data of (a) with real background noise (c) added. (f) Synthetic B-Scan data of (b) with real background noise (d) added.

### A. Model Testing

The performance results of the GPRInvNet on real data were validated in this study using a model testing process. A concrete experimental model with approximate dimensions of $4 \times 2 \times 0.7$ m (length × width × height) was built, as shown in Fig. 13. The middle section of the concrete model had not contained any defects and was referred to as a non-defect zone in this study. The remaining sections of the model were divided into several zones with approximate dimensions of $0.7 \times 0.7 \times 0.7$ m for different experimental processes. Rebars, cracks, and voids were deployed in different zones of the concrete model. A hollow acrylic box with the dimensions of $400 \times 600 \times 30$ mm was used in the concrete to represent the anhydrous cracks [Fig. 13(a)]. A waterproof plastic box measuring $400 \times 600 \times 200$ mm was utilized as a void [Fig. 13(b)], and filled with water to simulate water-bearing defects. Then, crack defects were deployed on the superficial layer at a depth of approximately 20 cm. The simulated water-bearing defects were placed on the bottom of the model. There were four rebars with diameters of 16 mm in the concrete model, which were located at intervals of approximately 15 cm. The distances between each of the defects were large enough to avoid interferences from two signals, as well as mitigate the impacts of any movements which may occur during the pouring of the concrete. Three of the defects were at depths of approximately 15 cm, and the remaining defect was located at a depth of 20 cm. The experiments were carried out 41 days after the concrete model had been formed.

This study utilized MALA Impulse Radar with a central frequency of 600 MHz in the experimental process to penetrate the concrete with a depth of 0.7 m, as shown in Fig. 13. The instrument integrated an antenna with its logger and transmitted the recorded data via WiFi to a tablet PC. The data were displayed in real-time in the format of the B-Scan images. The logged data could also be imported to a computer and further analyzed using professional software. In this article, the sampling point was set as 512 under a "wheel" mode, and the trace interval was set as 0.02 m.

### B. Data Processing

Due to the inhomogeneity of practical structures in which the detections were made, as well as the potential impact of noise under real environmental conditions, the detected real GPR B-Scan was more complicated than the synthetic data. Moreover, the training of DNN models required a large amount of data pairs, including the real B-Scans and the corresponding permittivity maps. It is difficult to acquire these data pairs in real engineering, and there currently is no available real GPR data set for training the DNN models. In order to generalize the GPRInvNet on real data, we employed the synthetic data integrated with real background noises for the purpose of data augmentation. The GPR B-Scan data from the non-defect zone of the concrete experimental model were logged as the background noise. Then, background noise of real data recorded from the concrete model were added into the synthetic B-Scan data for the purpose of training the GPRInvNet. Therefore, in the present study's experiment, not only the B-Scan data with defects in the concrete were recorded but also those without any defects. The B-Scan data containing defects were utilized only for testing purposes, and the B-Scan containing background noise was employed in the training process of the GPRInvNet.

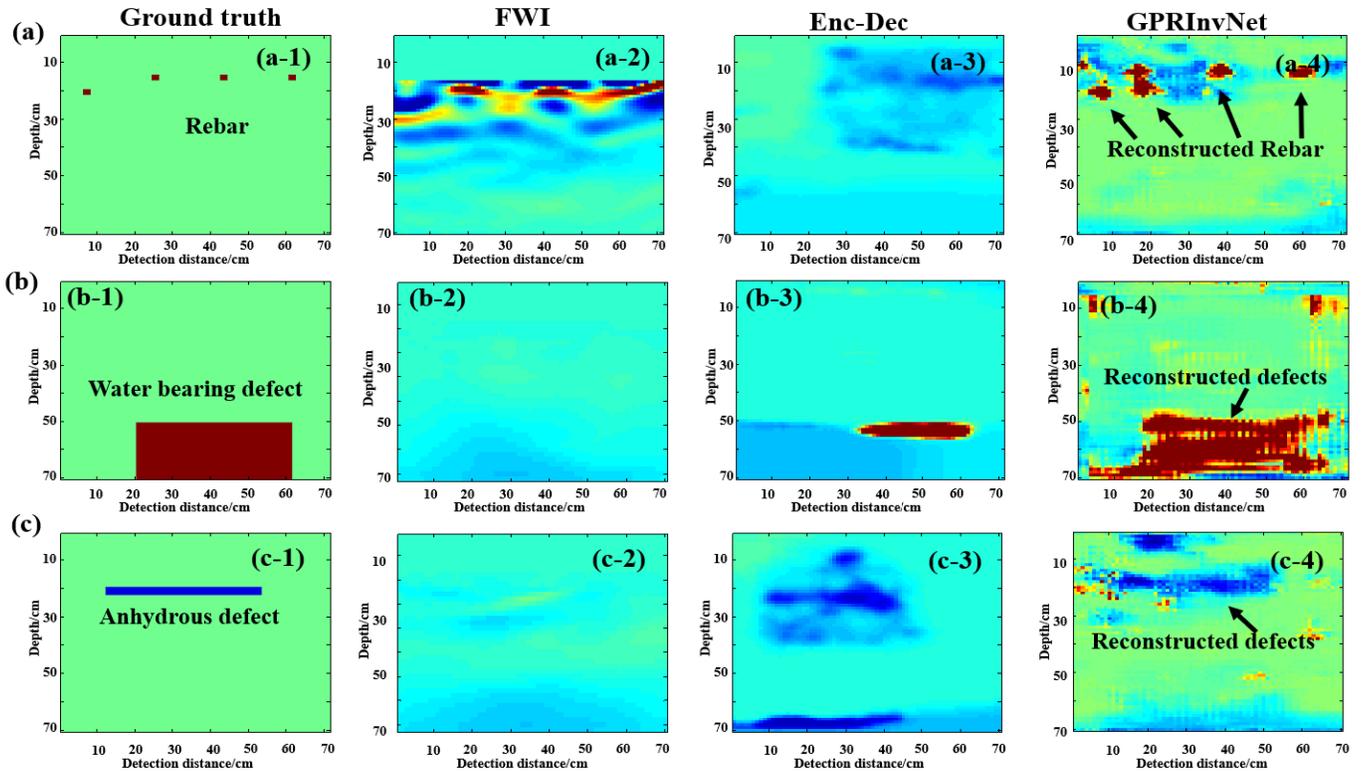

Fig. 16. Inversion results of the experimental data. (a) Ground truth and inversion results of the concrete model with rebars only. (b) Ground truth and inversion results of the concrete model with water-bearing defects. (c) Ground truth and inversion results of the concrete model with the anhydrous defect.

Prior to integrating the real data into the synthetic B-Scan data, preprocessing was performed on the real data. The preprocessing included several operations, such as static corrections, direct component removal, gain adjustments, background removal, filtering, and averaging. The B-Scan images for the rebars and defects following the preprocessing are shown in Fig. 14.

The real B-Scan data which contained only background noise were also pre-processed. The direct component of the background noise was removed and interpolation was employed to enlarge the time steps $T$ of the real background noise data to 800. Then, a sliding window method was used to randomly crop the background noise patches. A total of 187 background noise patches with dimensions of $800 \times 99$ ($T \times R$) were obtained at this step in the experimental process, as shown in Fig. 15(c) and (d). Amplitude normalization was then performed for both the background noise patches and the synthetic data. During the normalization process, the maximum amplitudes of the B-Scan images were multiplied by a weight coefficient which was randomly selected in the range of 0.5–2, in order to increase the diversity of the data. Finally, the data of the normalized background noise patches and synthetic B-Scan images were added per pixel to form a new B-Scan with real background noise. By doing so, a new data set which contained a large amount of data pairs was generated. Compared with the synthetic B-Scan data, the B-Scan in this new data set contained the information of background noises from the real environmental conditions. Therefore, the DNN models trained on this new data set may be applicable to the real GPR data inversion. Fig. 15 depicts some of the B-Scans in the new data set with background noises.

The GPRInvNet was retrained on the new data set for 100 epochs. Then, by adding the real background patches with synthetic data into the training and validation data sets, this study expected that the GPRInvNet would acquire new information, such as the inhomogeneity of the practical medium and the interference levels of noise in the real environment. It should be noted that there were no real data included in the new training data set, and the GPRInvNet was never trained using real data which depicted any hyperbolic echoes induced by the defects.

### C. Experimental Results and Analysis

The retrained GPRInvNet was tested using the real B-Scan data of the defects. The performance results of the GPRInvNet, Enc-Dec, and FWI were then compared. This study's comparative results pertaining to the different methods are shown in Fig. 16. The inversion results for the anhydrous cracks are presented in Fig. 16(a). Meanwhile, Fig. 16(b) and (c) depicts the inversion results for the rebars and water-bearing void, respectively.

As can be seen in Fig. 16, the overall performance results of the GPRInvNet were superior to those of the Enc-Dec and FWI, even for the real data. The shapes of the rebars, anhydrous cracks, and water-bearing voids which had been inverted using GPRInvNet were found to be almost identical to the ground truths. For example, the GPRInvNet had successfully recovered the locations and profiles of the four rebars [Fig. 16(a-4)]. On the contrary, the Enc-Dec had completely missed the rebar layer [Fig. 16(a-3)], and FWI was only able to determine the approximate distribution range of the rebars, rather than their shapes [Fig. 16(a-2)]. In addition, the water-bearing defect had well reconstructed by the GPRInvNet, as shown in Fig. 16(b-4). However, the Enc-Dec was only able to reconstruct a part of the defect, while the FWI had missed the water-bearing defect on the bottom of the model. In regard to the anhydrous cracks, all three methods were observed to be capable of inverting the profile of those defects, as can be seen in Fig. 16(c), but the shapes of the crack reconstructed by GPRInvNet were found to be closer to the true model. The GPRInvNet had obviously outperformed the other two methods for the rebars and water-bearing defects. Although the boundaries of the inverted defects were slightly blurred due to the inhomogeneity of the concrete in the experiment, the GPRInvNet was still able to provide rough profiles for those common types of defects.

It was determined through this study's experimental processes that the GPRInvNet had provided the best overall performance, especially in regard to the rebars and water-bearing defects. Moreover, the experimental results indicated that the GPRInvNet trained only on synthetic data could be effectively generalized on real GPR data by adding real background patches into the training data set.

## VI. Conclusion

In this article, a novel DNN-based architecture referred to as GPRInvNet was proposed for reconstructing high-quality relative permittivity maps of tunnel linings from GPR data. The GPRInvNet framework included a specially designed encoder for GPR data which was able to make full use of the GPR recordings, as well as retain the spatial alignments between the B-Scan images and the permittivity maps. Therefore, it was determined that the GPRInvNet had clear potential significance for improving the reconstruction of tunnel lining defects and assessing the status of complex defects in tunnel linings. Furthermore, the network was also found to have the potential to be employed in the inversion processes of GPR data in other GPR related applications, provided that sufficient training data were made available.

The GPRInvNet was first validated based on synthetic GPR data. Then, the GPRInvNet was successfully applied to reconstruct a permittivity map of tunnel linings containing complex defects with irregular geometric characteristics. It was determined to be capable of effectively reconstructing the dielectric properties and shapes of the most common types of tunnel lining defects, with clear boundaries observed. This study's comparative results demonstrated that the GPRInvNet had outperformed the existing baseline methods.

The performance results of the GPRInvNet were also verified through real GPR data experiments. A method was introduced to transfer GPRInvNet, which was first trained using only synthetic data, to real GPR data. The experimental results verified that the GPRInvNet was able to effectively invert the tunnel lining defects based on the real data, particularly the rebars and water-bearing defects. It was also revealed that the GPRInvNet had generalized the real GPR data by adding some background GPR acquisitions to the training data pool.

However, due to the complexity of the real GPR data, only rough profiles of the defects could be reconstructed. Also, this study did not perform validation experiments using field experimental data, and these issues will merit further research in the future.

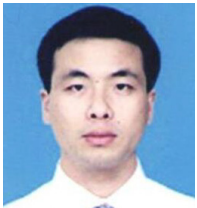

**Bin Liu** received the B.S. degree and the Ph.D. degree in civil engineering from Shandong University, Shandong, China, in 2005 and 2010, respectively.

He then joined the Geotechnical and Structural Engineering Research Center, Shandong University, where he is a Professor with the School of Qilu Transportation. His research area is engineering geophysical prospecting techniques, especially their applications in tunnels.

Dr. Liu is a member of Society of Exploration Geophysicists (SEG) and International Society for Rock Mechanics and Rock Engineering (ISRM). He serves as a Council Member for the Chinese Geophysical Society.

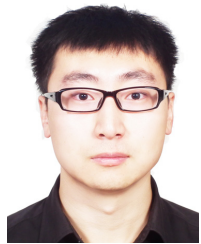

**Zhengfang Wang** received the Ph.D. degree in measurement technology and automatic instrument from Shandong University, Shandong, China, in 2014.

He is an Associate Professor with the School of Control Science and Engineering, Shandong University. His research interests include GPR data processing, detection and diagnosis of infrastructures and optic fiber sensors for infrastructures monitoring.

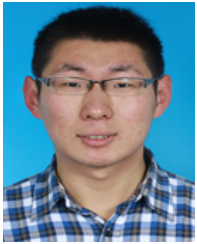

**Yuxiao Ren** received the bachelor's degree in mathematics from Shandong University, Shandong, China, 2014, and the master's degree in mathematics from Loughborough University, Loughborough, U.K., in 2015. He is pursuing the Ph.D. degree in civil engineering with Shandong University.

He joined Shandong University and has been studying for his doctoral degree in civil engineering since 2016. He is a Visiting Scholar with the Georgia Institute of Technology, under the supervision of Prof. F. Herrmann. His research interests include seismic modeling and imaging, full-waveform inversion and deep learning-based geophysical inversion.

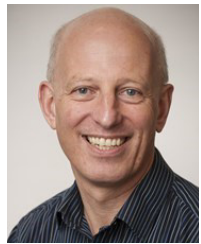

**Anthony G. Cohn** is a Full Professor with the School of Computing, University of Leeds, Leeds, U.K., and a Fellow with the Alan Turing Institute, U.K. He is a Fellow of the Royal Academy of Engineering, the Association for Advancement of Artificial Intelligence, and the European Association for Artificial Intelligence. His research interests are in artificial intelligence, knowledge representation and reasoning, cognitive vision, robotics, sensor fusion, and decision support systems. Since 2014, part of his research has focused on decision support systems for streetworks and utilities. The VAULT system which provides 24/7 real-time integrated utility data across Scotland arising from his Mapping the Underworld and VISTA projects won an IET Innovation Award and an NJUG Award for Avoiding Damage.

Mr. Cohn has received the Distinguished Service Awards from IJCAI and AAAI.

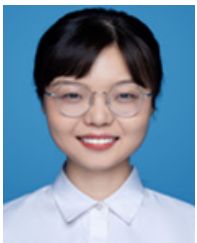

**Hanchi Liu** received the B.E. degree in information and electrical engineering from Shandong Jianzhu University, Shandong, China, in 2019. She is pursuing the master's degree with the School of Control Science and Engineering, Shandong University, Shandong, China.

Her research interests include defects recognition and deep learning-based geophysical inversion.

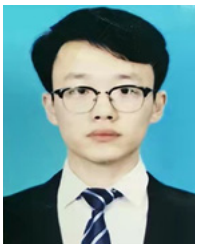

**Hui Xu** received the master's degree with Qilu Transportation, Shandong University, Shandong, China, in 2019.

He is working on 3-D object reconstruction.

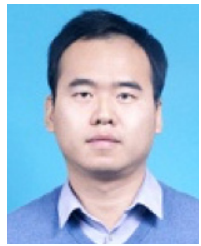

**Peng Jiang** (Member, IEEE) received the B.S. and Ph.D. degrees in computer science and technology from Shandong University, Shandong, China, in 2010 and 2016, respectively.

He is a Research Assistant with the School of Qilu Transportation, Shandong University. He has authored or coauthored many works on top-tier venues, including ICCV, NeurIPS(NIPS), the IEEE TRANSACTIONS ON IMAGE PROCESSING (TIP), and the IEEE TRANSACTIONS ON GEOSCIENCE AND REMOTE SENSING (TGRS). He is focusing on deep learning-based geophysical inversion. His research spans various areas, including computer vision, image processing, machine learning, and deep learning.